\documentclass[sigconf, nonacm]{acmart}

\usepackage{graphicx}
\usepackage{balance}  
\usepackage[linesnumbered,ruled,noend]{algorithm2e}
\usepackage{graphicx, caption, subcaption}
\usepackage{multirow}
\usepackage{booktabs}
\usepackage{color}
\usepackage{hyperref}
\usepackage{amsmath}
\usepackage{cancel}
\usepackage{ifthen}

\newcommand\vldbdoi{XX.XX/XXX.XX}
\newcommand\vldbpages{XXX-XXX}
\newcommand\vldbvolume{14}
\newcommand\vldbissue{1}
\newcommand\vldbyear{2020}
\newcommand\vldbauthors{\authors}
\newcommand\vldbtitle{\shorttitle} 
\newcommand{\ig}{Inspector Gadget}
\newcommand{\snorkel}{Snorkel}
\newcommand{\snuba}{Snuba}
\newcommand{\goggles}{GOGGLES}

\newcommand{\squishlist}{ 
   \begin{list}{$\bullet$}
    { \setlength{\itemsep}{0pt}      \setlength{\parsep}{3pt} 
      \setlength{\topsep}{3pt}       \setlength{\partopsep}{0pt}
      \setlength{\leftmargin}{1.5em} \setlength{\labelwidth}{1em}
      \setlength{\labelsep}{0.5em} } }

\newcommand{\squishend}{
    \end{list}  } 

\newboolean{techreport}
\setboolean{techreport}{true}
\newboolean{color}
\setboolean{color}{false}

\ifthenelse{\boolean{color}}{
\newcommand{\rv}[1]{\textcolor{blue}{#1}}
\newcommand{\tr}[1]{\textcolor{red}{#1}}
}{
\newcommand{\rv}[1]{\textcolor{black}{#1}}
\newcommand{\tr}[1]{\textcolor{black}{#1}}
}

\begin{document}

\ifthenelse{\boolean{techreport}}{
\title{Inspector Gadget: A Data Programming-based Labeling System for Industrial Images}
}
{
\title{Inspector Gadget: A Data Programming-based Labeling System for Industrial Images [Scalable Data Science]}
}


\author{Geon Heo, Yuji Roh, Seonghyeon Hwang, Dayun Lee, Steven Euijong Whang}
\affiliation{%
  \institution{Korea Advanced Institute of Science and Technology}
  \city{Daejeon}
  \country{Republic of Korea}
}
\email{{geon.heo, yuji.roh, sh.hwang, dayun.lee, swhang}@kaist.ac.kr}

\begin{abstract}
As machine learning for images becomes democratized in the Software 2.0 era, one of the serious bottlenecks is securing enough labeled data for training. This problem is especially critical in a manufacturing setting where smart factories rely on machine learning for product quality control by analyzing industrial images. Such images are typically large and may only need to be partially analyzed where only a small portion is problematic (e.g., identifying defects on a surface). Since manual labeling these images is expensive, weak supervision is an attractive alternative where the idea is to generate weak labels that are not perfect, but can be produced at scale. Data programming is a recent paradigm in this category where it uses human knowledge in the form of labeling functions and combines them into a generative model. Data programming has been successful in applications based on text or structured data and can also be applied to images usually if one can find a way to convert them into structured data. In this work, we expand the horizon of data programming by directly applying it to images without this conversion, which is a common scenario for industrial applications. We propose \ig{}, an image labeling system that combines crowdsourcing, data augmentation, and data programming to produce weak labels at scale for image classification. \rv{We perform experiments on real industrial image datasets and show that \ig{} obtains better performance than other weak-labeling techniques: \snuba{}, \goggles{}, and self-learning baselines using convolutional neural networks (CNNs) without pre-training.}

\end{abstract}

\maketitle

\ifthenelse{\boolean{techreport}}{}
{
\vspace{-0.2cm}
\begingroup\small\noindent\raggedright\textbf{PVLDB Reference Format:}\\
\vldbauthors. \vldbtitle. PVLDB, \vldbvolume(\vldbissue): \vldbpages, \vldbyear.\\
\href{https://doi.org/\vldbdoi}{doi:\vldbdoi}
\endgroup
\begingroup
\renewcommand\thefootnote{}\footnote{\noindent
This work is licensed under the Creative Commons BY-NC-ND 4.0 International License. Visit \url{https://creativecommons.org/licenses/by-nc-nd/4.0/} to view a copy of this license. For any use beyond those covered by this license, obtain permission by emailing \href{mailto:info@vldb.org}{info@vldb.org}. Copyright is held by the owner/author(s). Publication rights licensed to the VLDB Endowment. \\
\raggedright Proceedings of the VLDB Endowment, Vol. \vldbvolume, No. \vldbissue\ %
ISSN 2150-8097. \\
\href{https://doi.org/\vldbdoi}{doi:\vldbdoi} \\
}\addtocounter{footnote}{-1}\endgroup
}

\section{Introduction}
\label{sec:introduction}

In the era of Software 2.0, machine learning techniques for images are becoming democratized where the applications range from manufacturing to medical. For example, smart factories regularly use computer vision techniques to classify defective and non-defective product images~\cite{oztemel2018literature}. In medical applications, MRI scans are analyzed to identify diseases like cancer~\cite{47793}. However, many companies are still reluctant to adapt machine learning due to the lack of labeled data where manual labeling is simply too expensive~\cite{DBLP:journals/debu/Stonebraker19}.

\begin{figure}[t]
\centering
  \includegraphics[width=0.9\columnwidth]{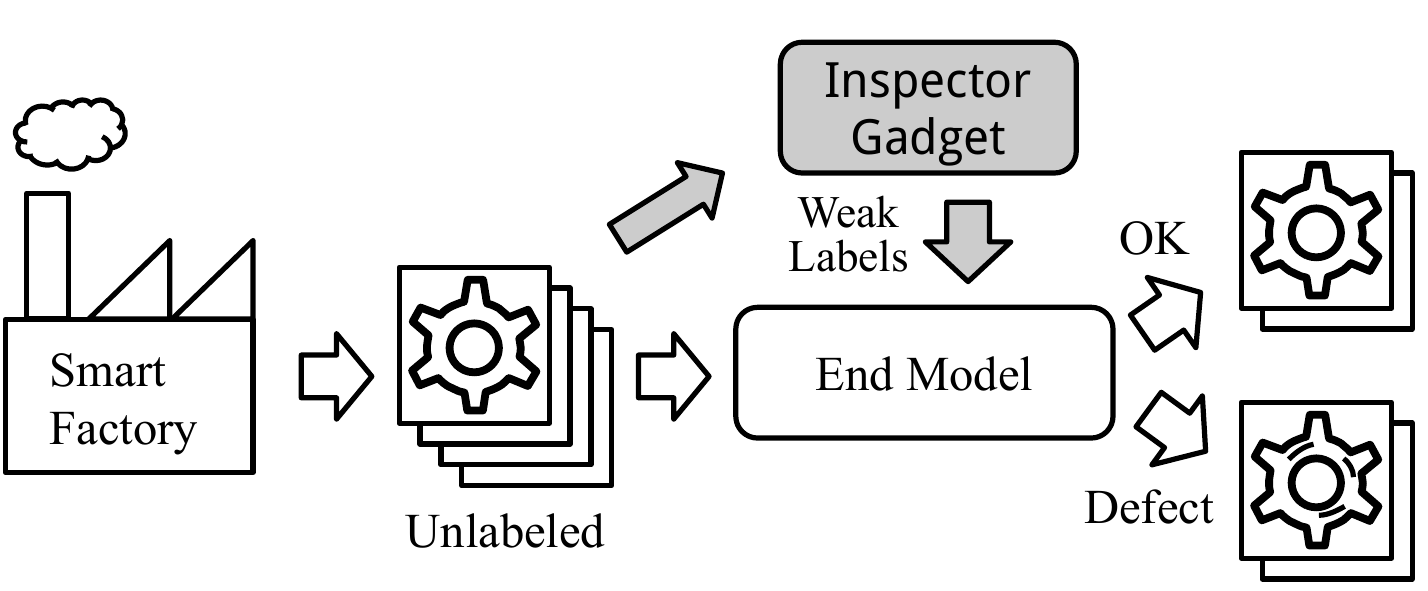}
  \vspace{-0.2cm}
  \caption{Labeling industrial images with \ig{} in a smart factory application.}
  \label{fig:scenario}
  \vspace{-0.3cm}
\end{figure}

We focus on the problem of scalable labeling for classification where {\em large} images are {\em partially analyzed}, and there are {\em few or no labels} to start with. 
Although many companies face this problem, it has not been studied enough. Based on a collaboration with a large manufacturing company, we provide the following running example. Suppose there is a smart factory application where product images are analyzed for quality control (Figure~\ref{fig:scenario}). \rv{These images taken from industrial cameras usually have high-resolution.} 
 The goal is to look at each image and tell if there are certain defects (e.g., identify scratches, bubbles, and stampings). For convenience, we hereafter use the term {\em defect} to mean a part of an image of interest.

A conventional solution is to collect enough labels manually and train say a convolutional neural network on the training data. However, fully relying on crowdsourcing for image labeling can be too expensive. In our application, we have heard of domain experts demanding six-figure salaries, which makes it infeasible to simply ask them to label images. In addition, relying on general crowdsourcing platforms like Amazon Mechanical Turk may not guarantee high-enough labeling quality.

Among the possible methods for data labeling (see an extensive survey~\cite{DBLP:journals/tkde/RohHW19}), 
\rv{weak supervision is an important branch of research where the idea is to semi-automatically generate labels that are not perfect like manual ones. Thus, these generated labels are called {\em weak labels}, but they have reasonable quality where the quantity compensates for the quality. Data programming~\cite{DBLP:conf/nips/RatnerSWSR16} is a representative weak supervision technique of employing humans to develop labeling functions (LFs) that individually perform labeling (e.g., identify a person riding a bike), perhaps not accurately. However, the combination of inaccurate LFs into a generative model results in probabilistic labels with reasonable quality.} These weak labels can then be used to train an end discriminative model.


So far, data programming has been shown to be effective in finding various relationships in text and structured data~\cite{Ratner:2017:SFT:3035918.3056442}. Data programming has also been successfully applied to images where they are usually converted to structured data beforehand~\cite{DBLP:conf/nips/VarmaHBKBRR17,DBLP:conf/icml/VarmaSHRR19}. 
However, this conversion limits the applicability of data programming.
As an alternative approach, \goggles{}~\cite{DBLP:journals/corr/abs-1903-04552} demonstrates that, on images, automatic approaches using pre-trained models may be more effective. Here the idea is to extract semantic prototypes of images using the pre-trained model and then cluster and label the images using the prototypes. However, \goggles{} also has limitations (see Section~\ref{sec:accuracyexperiments}), and it is not clear if it is the only solution for generating training data for image classification.

We thus propose \ig{}, which opens up a new class of problems for data programming by enabling direct image labeling at scale without the need to convert to structured data using a combination of crowdsourcing, data augmentation, and data programming techniques. \ig{} provides a crowdsourcing workflow where workers identify {\em patterns} that indicate defects. Here we make the tasks easy enough for non-experts to contribute. These patterns are augmented using general adversarial networks (GANs)~\cite{DBLP:conf/nips/GoodfellowPMXWOCB14} and policies~\cite{DBLP:journals/corr/abs-1805-09501}. Each pattern effectively becomes a labeling function by being matched with other images. The similarities are then used as features to train a multi-layer perceptron (MLP), which generates weak labels. 

In our experiments, \ig{} performs better overall than state-of-the-art methods: \snuba{}, \goggles{}, and self-learning baselines that use CNNs (VGG-19~\cite{DBLP:journals/corr/SimonyanZ14a} and MobileNetV2~\cite{DBLP:conf/cvpr/SandlerHZZC18}) without pre-training. We release our code as a community resource~\cite{github}.

In the rest of the paper, we present the following:
\squishlist
    \item The architecture of \ig{} (Section~\ref{sec:overview}).
    \item The component details of \ig{}:
    \squishlist
        \item Crowdsourcing workflow for helping workers identify patterns (Section~\ref{sec:crowdsourcing}).
        \item Pattern augmenter for expanding the patterns using GANs and policies (Section~\ref{sec:dataaugmentation}).
        \item Feature generator and labeler for generating similarity features and producing weak labels (Section~\ref{sec:trainer}).
    \squishend
    \item Experimental results where \ig{} outperforms \rv{other image labeling techniques} -- \snuba{}, \goggles{}, and self-learning baselines using CNNs -- where there are few or no labels to start with (Section~\ref{sec:experiments}).
\squishend

\section{Overview}
\label{sec:overview}

\begin{figure}[t]
  \includegraphics[width=0.9\columnwidth]{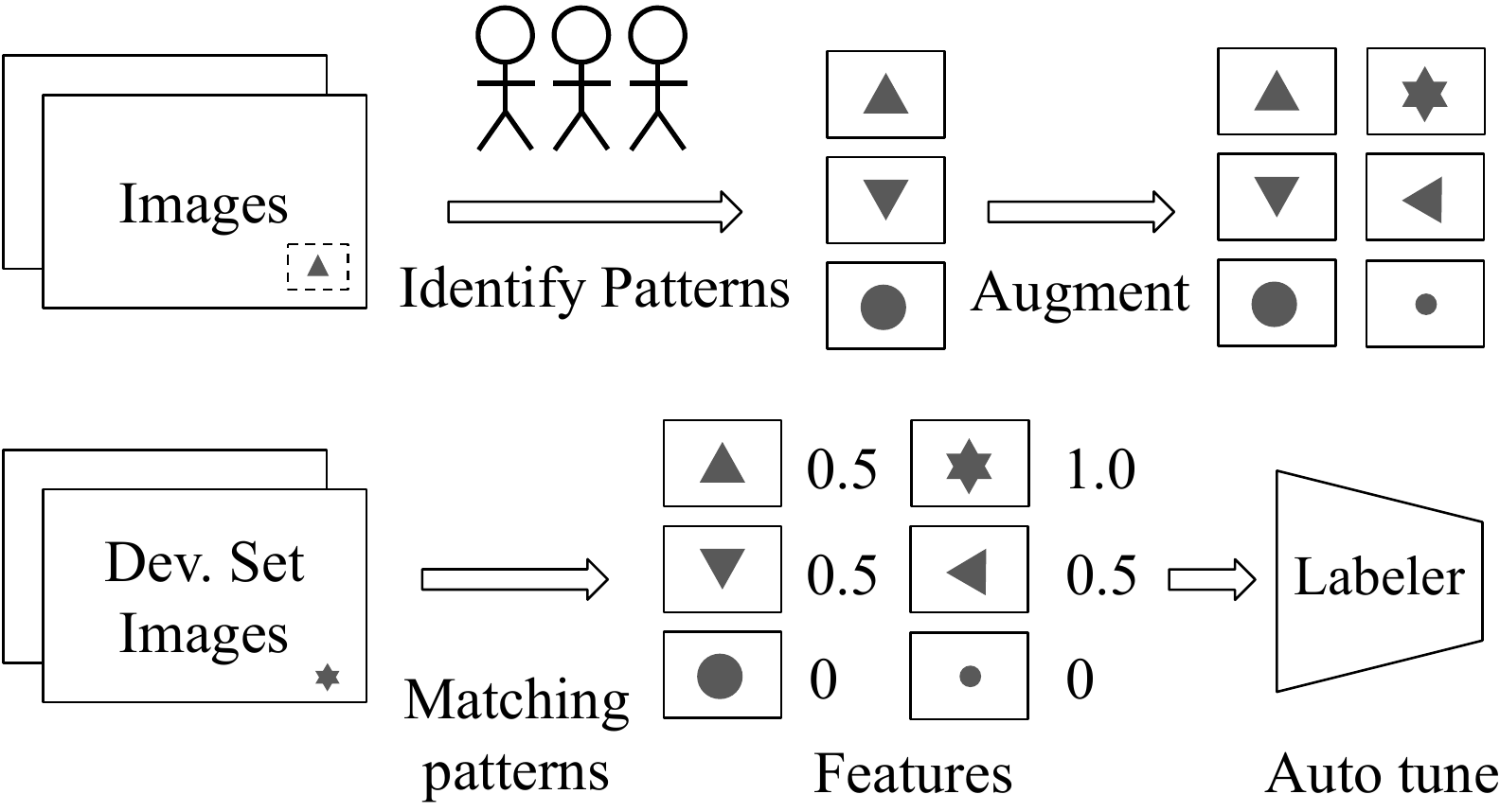}
  \caption{An overview of how \ig{} constructs a model (labeler) that generates weak labels.}
\label{fig:overview}
\vspace{-0.1cm}
\end{figure}

\begin{figure}[t]
  \includegraphics[width=\columnwidth]{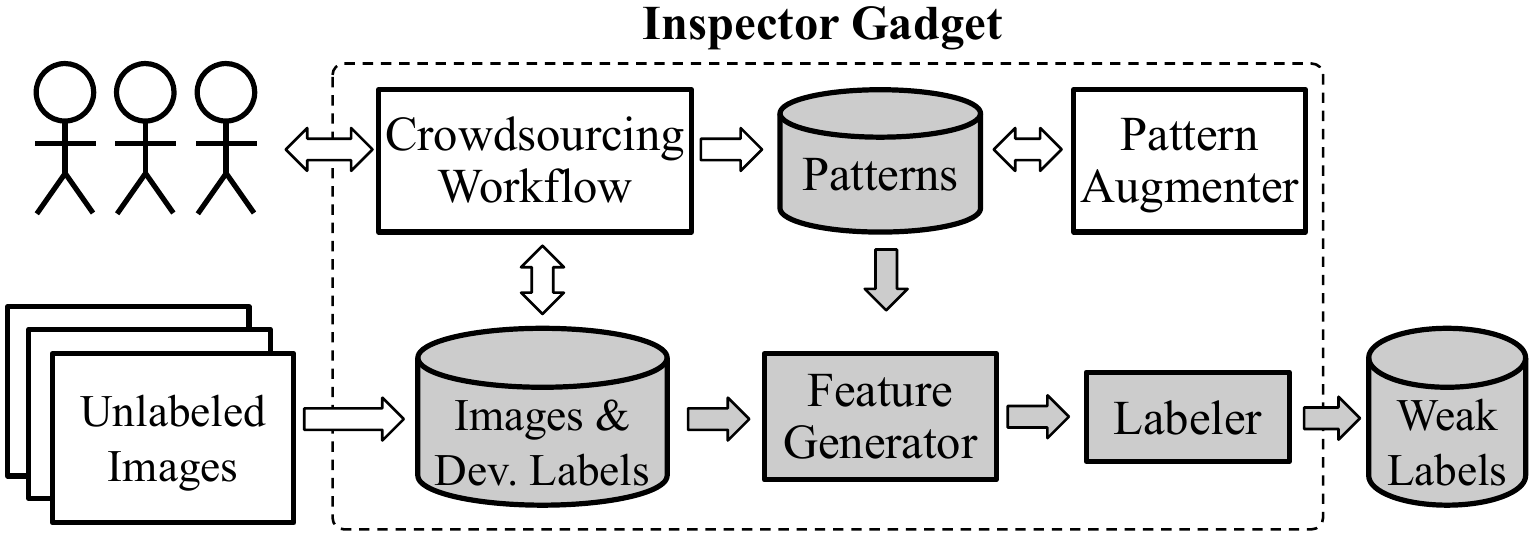}
  \caption{The architecture of \ig{}.}
  \label{fig:architecture}
  \vspace{-0.2cm}
\end{figure}

 The main technical contribution of \ig{} is its effective combination of crowdsourcing, data augmentation, and data programming for scalable image labeling for classification. Figure~\ref{fig:overview} shows the overall process of \ig{}. First, a crowdsourcing workflow helps workers identify patterns of interest from images that may indicate defects. While the patterns are informative, they may not be enough and are thus augmented using generative adversarial networks (GANs)~\cite{DBLP:conf/nips/GoodfellowPMXWOCB14} and policies~\cite{DBLP:conf/cvpr/CubukZMVL19}. Each pattern effectively becomes a labeling function where it is compared with other images to produce similarities that indicate whether the images contain defects. A separate development set is used to train a small model that uses the similarity outputs as features. This model is then used to generate \rv{weak labels of images indicating the defect types in the test set.} Figure~\ref{fig:architecture} shows the architecture of the \ig{} system. After training the Labeler, \ig{} only utilizes the components highlighted in gray for generating weak labels. In the following sections, we describe each component in more detail.

\begin{figure*}[t]
  \center
     \includegraphics[width=0.85\textwidth]{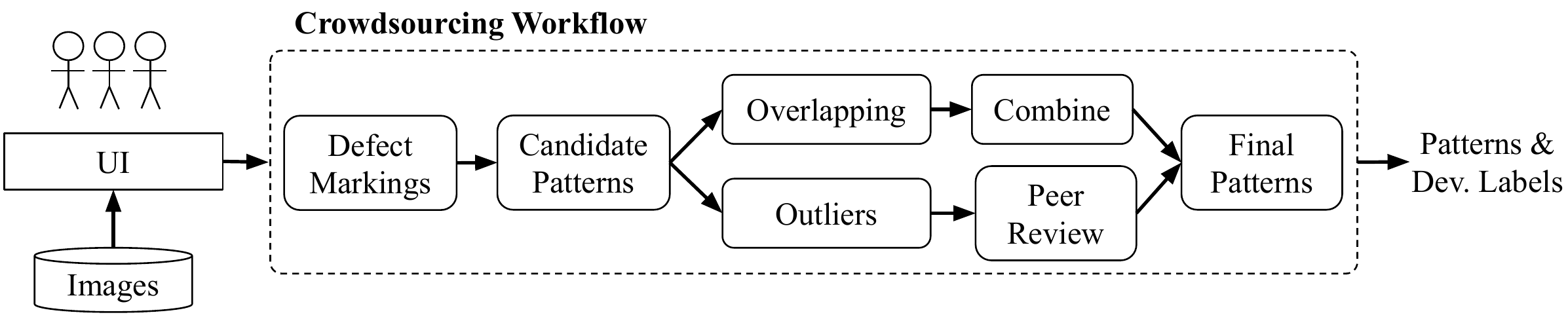}
     \caption{The crowdsourcing workflow of \ig{}. Crowd workers can interact with a UI and provide bounding boxes that identify defects. The boxes are used as patterns, which are combined or go through a peer review phase.}
     \vspace{-0.2cm}
 \label{fig:crowdsourcingworkflow}
 \vspace{-0.2cm}
\end{figure*}

\section{Crowdsourcing Workflow}
\label{sec:crowdsourcing}

\begin{figure}[t]
  \centering
     \includegraphics[width=0.9\columnwidth]{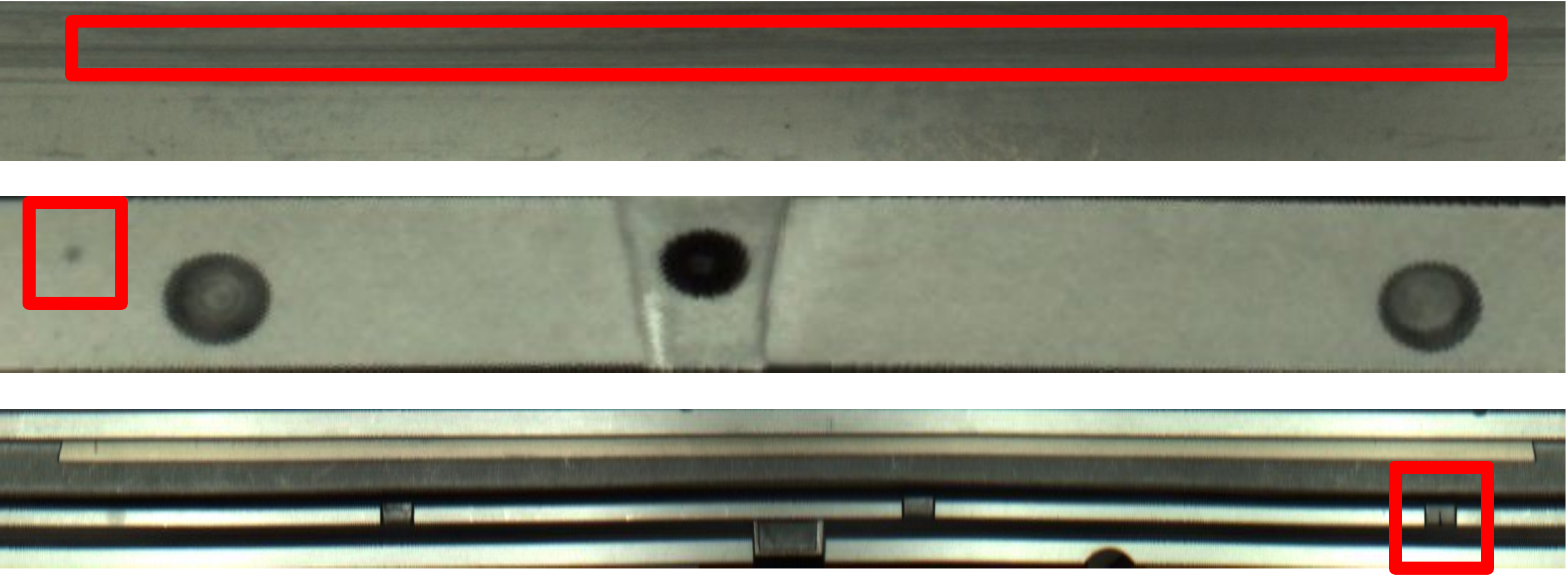}
     \vspace{-0cm}
\caption{Sample images in the {\sf Product} dataset (see Section~\ref{sec:settings}) containing scratch, bubble, and stamping defects where we highlight each defect with a bounding box. }%
\label{fig:productdata}
\vspace{-0.5cm}
\end{figure}

Since collecting enough labels on the entire images are too expensive, we would like to utilize human knowledge as much as possible and reduce the additional amount of labeled data. We propose a crowdsourcing workflow shown in Figure~\ref{fig:crowdsourcingworkflow}. First, the workers mark defects using bounding boxes through a UI. Since the workers are not necessarily experts, the UI educates them how to identify defects beforehand. The bounding boxes in turn become the patterns we use to find other defects. Figure~\ref{fig:productdata} shows sample images of a real-world smart factory dataset (called {\sf Product}; see Section~\ref{sec:settings} for a description) where defects are highlighted with red boxes. \rv{Notice that the defects are not easy to find as they are small and mixed with other parts of the product.} 


As with any crowdsourcing application, we may run into quality control issues where the bounding boxes of the workers vary even for the same defect. \ig{} addresses this problem by first combining overlapping bounding boxes together. 
\ifthenelse{\boolean{techreport}}
{\tr{While there are several ways to combine boxes, we find that averaging their coordinates works reasonably well. The two other strategies we considered were to take the ``union'' of coordinates (i.e., find the coordinates that cover the overlapping boxes), or take the ``intersection'' of coordinates (i.e., find the coordinates for the common parts of the boxes). However, the union strategy tends to generate patterns that are too large, while the intersection strategy has the opposite problem of generating tiny patterns. Hence, we only use the average strategy in our experiments.}}
{While there are several ways to combine boxes \rv{(see our technical report~\cite{inspectorgadgettr} for details)}, we find that averaging their coordinates works reasonably well. }
For the remaining outlier boxes, \ig{} goes through a peer review phase where workers discuss which ones really contain defects. In Section~\ref{sec:workflowexperiments}, we perform ablation tests to show how each of these steps helps improve the quality of patterns. 

Another challenge is determining how many images must be annotated to generate enough patterns. In general, we may not have statistics on the portion of images that have defects. Hence, our solution is to randomly select images and annotate them until the number of defective images exceeds a given threshold. In our experiments, identifying tens of defective images is sufficient (see the $N_V^D$ values in Table~\ref{tbl:datasets}). All the annotated images form a {\em development set}, which we use in later steps \rv{for training the labeler.}

\rv{The crowdsourcing workflow can possibly be automated using pre-trained region proposal networks (RPNs)~\cite{ren2015faster}. However, this approach requires other training data with similar defects and bounding boxes, which seldom exist for our applications.}



\section{Pattern Augmenter}
\label{sec:dataaugmentation}

Pattern augmentation is a way to compensate for the possible lack of patterns even after using crowdsourcing. The patterns can be lacking if not enough human work is done to identify all possible patterns and especially if there are not enough images containing defects \rv{(i.e., there is a class imbalance)} so one has to go through many images just to encounter a negative label. We would thus like to automatically generate more patterns without resorting to more expensive crowdsourcing. 

We consider two types of augmentation -- GAN-based~\cite{DBLP:conf/nips/GoodfellowPMXWOCB14} and policy-based~\cite{DBLP:journals/corr/abs-1805-09501}. The two methods complement each other and have been used together in medical applications for identifying lesions~\cite{DBLP:journals/ijon/Frid-AdarDKAGG18}. GAN-based augmentation is good at generating random variations of existing defects that do not deviate significantly. On the other hand, policy-based augmentation is better for generating specific variations of defects that can be quite different, exploiting domain knowledge. In Section~\ref{sec:dataaugmentationexperiments}, we show that neither augmentation subsumes the other, and the two can be used together to produce the best results.

The augmentation can be done efficiently because we are augmenting small patterns instead of the entire images. For high-resolution images, it is sometimes infeasible to train a GAN at all. In addition, if most of the image is not of interest for analysis, it is difficult to generate fake parts of interest while leaving the rest of the image as is. By only focusing on augmenting small patterns, it becomes practical to apply sophisticated augmentation techniques.

\subsection{GAN-based Augmentation}
\label{subsec:ganaugmentation}
The first method is to use generative adversarial networks (GAN) to generate variations of patterns that are similar to the existing ones. Intuitively, these augmented patterns can fill in the gaps of the existing patterns. The original GAN~\cite{DBLP:conf/nips/GoodfellowPMXWOCB14} trains a generator to produce realistic fake data where the goal is to deceive a discriminator that tries to distinguish the real and fake data. More recently, many variations have been proposed (see a recent survey~\cite{DBLP:journals/corr/abs-1906-01529}). 


\rv{We use a Relativistic GAN (RGAN)~\cite{DBLP:conf/iclr/Jolicoeur-Martineau19}, which can efficiently generate more realistic patterns than the original GAN.} The formulation of RGAN is:
\begin{align*}
\max_D \mathbb{E}_{(x_r,G(z))\sim (\mathbb{P}, \mathbb{Q})}[\mathrm{log}(\sigma(D(x_r)-D(G(z))))]
\\
\max_G \mathbb{E}_{(x_r,G(z))\sim (\mathbb{P}, \mathbb{Q})}[\mathrm{log}(\sigma(D(G(z))-D(x_r)))]
\end{align*}
where $G$ is the generator, $x_r$ is real data, $D(x)$ is the probability that $x$ is real, $z$ is a random noise vector that is used by $G$ to generate various fake data, and $\sigma$ is the sigmoid function. 
\rv{While training, the discriminator of RGAN not only distinguishes data, but also tries to maximize the difference between two probabilities: the probability that a real image is actually real, and the probability that a fake image is actually fake. This setup enforces fake images to be more realistic, in addition to simply being distinguishable from real images as in the original GAN.}
We also use Spectral Normalization~\cite{DBLP:journals/corr/abs-1802-05957}, which is a commonly-used technique applied to a neural network structure where the discriminator restricts the gradient to adjust the training speed for better training stability. 

\begin{figure}[t]
  \centering
     \includegraphics[width=0.8\columnwidth]{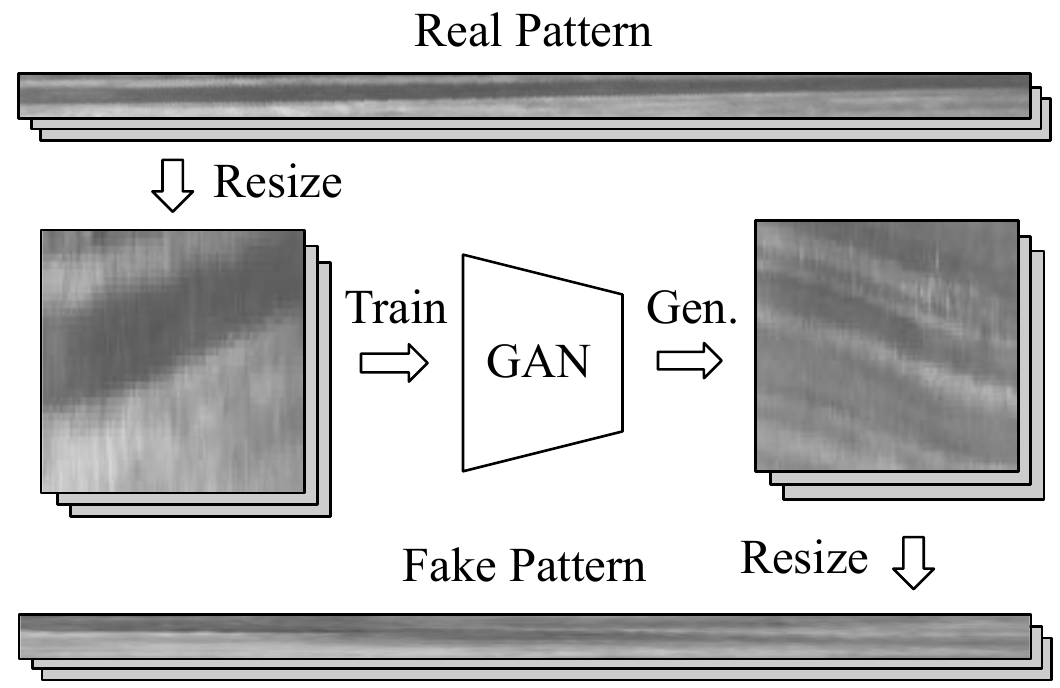} 
     \caption{GAN-based augmentation on a pattern containing a scratch defect from the {\sf Product} dataset.}
 \label{fig:gan_process}
 \vspace{-0.2cm}
\end{figure}

\rv{Another issue is that neural networks assume a fixed size and shape of its inputs. While this assumption is reasonable for a homogeneous set of images, the patterns may have different shapes. We thus fit patterns to a fixed-sized square shape by resizing them before augmentation.} Then we re-adjust new patterns into \rv{one of the original sizes in order to properly find other defects.} Figure~\ref{fig:gan_process} shows the entire process applied to an image. \rv{In Section~\ref{sec:dataaugmentationexperiments}, we show that the re-sizing of patterns is effective in practice.}

\subsection{Policy-based Augmentation}

\begin{figure}[t]
  \centering
     \includegraphics[width=0.9\columnwidth]{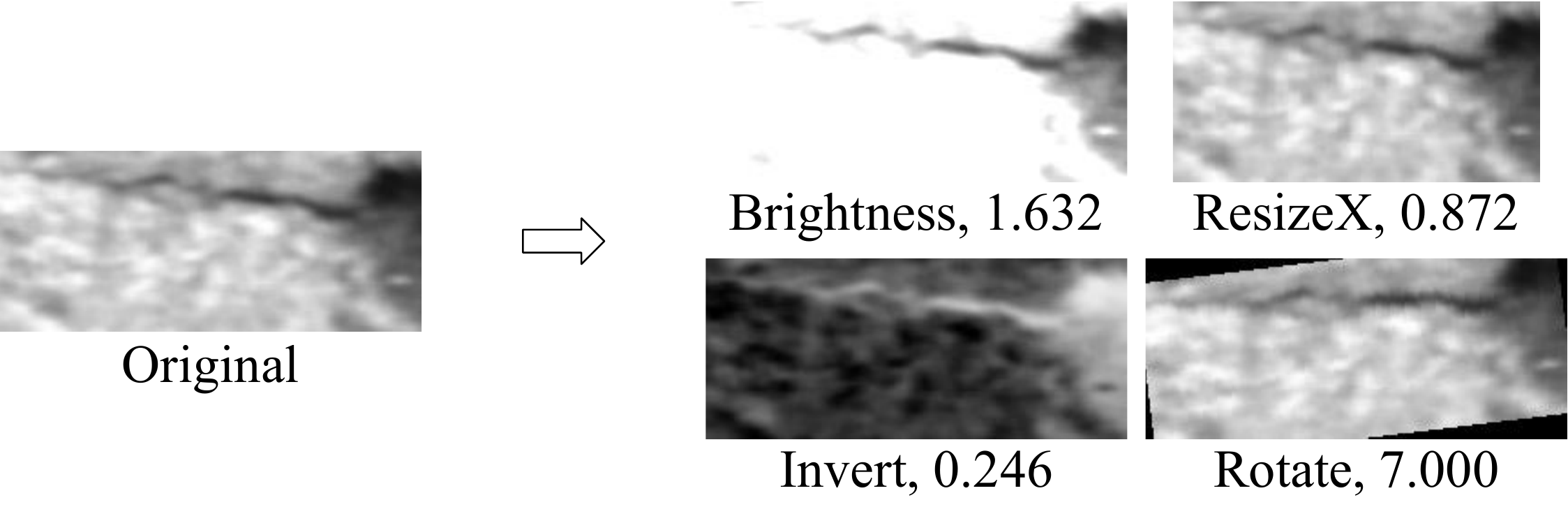} 
     \caption{Policy-based augmentation on a pattern with a crack defect from the {\sf KSDD} dataset (see Section~\ref{sec:settings}). For each augmentation, we show the operation and magnitude.}
 \label{fig:augmentation-policy}
 \vspace{-0.3cm}
\end{figure}

Policies~\cite{DBLP:journals/corr/abs-1805-09501} have been proposed as another way to augment images and complement the GAN approach. The idea is to use manual policies to decide how exactly an image is varied. 
\rv{Figure~\ref{fig:augmentation-policy} shows the results of applying four example policies on a surface defect from the {\sf KSDD} dataset (see description in Section~\ref{sec:settings}). 
}
Policy-based augmentation is effective for patterns where applying  operations based on human knowledge may result in quite different, but valid patterns. For example, if a defect is line-shaped, then it makes sense to stretch or rotate it. There are two parameters to configure: the operation to use and the magnitude of applying that operation. Recently, policy-based techniques have become more automated, e.g., AutoAugment~\cite{DBLP:journals/corr/abs-1805-09501} uses reinforcement learning to decide to what extent can policies be applied together.

\rv{We use an simpler approach than AutoAugment.} Among certain combinations of policies, we choose the ones that work best on the development set. We first split the development set into train and test sets. For each policy, we specify a range for the magnitudes and choose 10 random values within that range. We then iterate all combinations of three policies. For each combination, we augment the patterns in the train set using the 10 magnitudes and train a model (see details in Section~\ref{sec:trainer}) on the train set images until convergence. Then we evaluate the model on the separate test set. Finally, we use the policy combination that results in the best accuracy and apply it to the entire set of patterns.


\section{Weak Label Generation}
\label{sec:trainer}

Once \ig{} gathers patterns and augments them, the next step is to generate features of images (also called primitives according to \snuba{}~\cite{DBLP:journals/pvldb/VarmaR18}) and train a model that can produce weak labels. Note that the way we generate features of images is more direct than existing data programming systems that first convert images to structured data \rv{with interpretable features using object detection} (e.g., identify a vehicle) before applying labeling functions.

 
\subsection{Feature Generator}
\label{sec:featuregeneration}

\rv{\ig{} provides feature generation functions (FGFs) that match all generated patterns with the new input image to identify similar defects on any location and return the similarities with the patterns. 
Notice that the output of an FGF is different than a conventional labeling function in data programming where the latter returns a weak label per image.    
A vector that consists of all output values of the FGFs on each image is used as the input of the labeler.}
Depending on the type of defect, the pattern matching may differ. A na\"ive approach is to do an exact pixel-by-pixel comparison, but this is unrealistic when the defects have variations. Instead, a better way is to compare the distributions of pixel values. This comparison is more robust to slight variations of a pattern. On the other hand, there may be false positives where an obviously different defect matches just because its pixels have similar distributions. In our experiments, we found that comparing distributions on the x and y axes using normalized cross correlation~\cite{opencv} is effective in reducing such false positives.
Given an image $I$ with pixel dimensions $W\times H$ and pattern $P_i$ with dimensions $w \times h$, \rv{the $i^{th}$ FGF $f_i$} is defined as: 

\vspace{-0.1cm}
\begin{align*}
\rv{f_i(I)} = \max_{x,y}\left |\frac{\sum _{x',y'} P_i(x',y')\cdot I(x+x',y+y')}{\sqrt{\sum _{x',y'} P_i(x',y')^{2}\cdot \rv{\sum _{x',y'}} I(x+x',y+y')^{2}}}\right |
\end{align*}

where $0 \leq x < W-w$, $0 \leq y < H-h$, $0 \leq x' < w$, and $0 \leq y' < h$. When matching a pattern against an image, a straightforward approach is to make a comparison with every possible region of the same size in the image. However, scanning the entire image may be too time consuming. Instead, we use a pyramid method~\cite{adelson1984pmi} where we first search for candidate parts of an image by reducing the resolutions of the image and pattern and performing a comparison quickly. Then just for the candidates, we perform a comparison using the full resolution.

\begin{table*}[t]
  \caption{For each of the five datasets, we show the image size, the dataset size($N$) and number of defective images($N^D$), the development set size($N_V$) and number of defective images within it($N_V^D$), the types of defects detected, and the task type. }
  \vspace{-0.1cm}
  \begin{tabular}{cccccc}
    \toprule
    {\bf Dataset} & {\bf Image size} & {\bf $N$ ($N^D$)} & {\bf $N_{V}$ ($N_V^D$)} & {\bf Defect Type} & {\bf Task Type}\\
    \midrule
    \multirow{1}{*}{\sf KSDD}~\cite{Tabernik2019JIM} & \multirow{1}{*}{500 x 1257} & \multirow{1}{*}{399 (52)} & 78 (10) & Crack & Binary\\
    \midrule
    \multirow{1}{*}{{\sf Product (scratch)}} & \multirow{1}{*}{162 x 2702} & \multirow{1}{*}{1673 (727)} & 170 (76) & Scratch & Binary\\
    \midrule
    \multirow{1}{*}{{\sf Product (bubble)}} & \multirow{1}{*}{77 x 1389} & \multirow{1}{*}{1048 (102)} & 104 (10) & Bubble & Binary\\
    \midrule
    \multirow{1}{*}{{\sf Product (stamping)}} & \multirow{1}{*}{161 x 5278} & \multirow{1}{*}{1094 (148)} & 109 (15) & Stamping & Binary\\
    \midrule
    \multirow{2}{*}{{\sf NEU}~\cite{He2019TIM}} & \multirow{2}{*}{200 x 200} & \multirow{2}{*}{300 per defect} & \multirow{2}{*}{100 per defect} & Rolled-in scale, Patches, Crazing, & \multirow{2}{*}{Multi-class}\\ 
    & & & & Pitted surface, Inclusion, Scratch & \\
    \bottomrule
  \end{tabular}
  \label{tbl:datasets}
 \vspace{-0.3cm}
\end{table*}

\subsection{Labeler}
\label{sec:modeltraining}

After the features are generated, \ig{} trains a model on the output similarities of the FGFs, where the goal is to produce weak labels. 
The model can have any architecture and be small because there are not as many features as say the number of pixels in an image. We use a multilayer perceptron (MLP) because it is simple, but also has good performance compared to other models. An interesting observation is that, depending on the model architecture (e.g., the number of layers in an MLP), the model accuracy can vary significantly as we demonstrate in Section~\ref{sec:modeltuningexperiments}. \ig{} thus performs model tuning where it chooses the architecture that has the best accuracy results on the development set. This feature is powerful compared to existing CNN approaches where the architecture is complicated, and it is too expensive to consider other variations.


\rv{The labeling process after training the labeler consists of two steps.} \rv{First, the patterns are matched to unlabeled images for generating the features. }
Second, the trained labeler is applied on the features to make a prediction \rv{for each unlabeled image.} We note that latency is not the most critical issue because we are generating weak labels, which are used to construct the training data for the end discriminative model. Training data construction is usually done in batch mode instead of say real time.

\section{Experiments}
\label{sec:experiments}

We evaluate \ig{} on real datasets and answer the following questions.

\squishlist
    \item How accurate are the weak labels of \ig{} compared to \rv{other labeling methods} and are they useful when training the end discriminative model?
    \item How useful is each component of \ig{}? 
    \item What are the errors made by \ig{}?
\squishend

We implement \ig{} in Python \rv{and use the OpenCV library} and three machine learning libraries: Pytorch, TensorFlow, and Scikit-learn. We use an Intel Xeon CPU to train our MLP models and an NVidia Titan RTX GPU to train larger CNN models. Other details can be found in our released code~\cite{github}.

\subsection{Settings}
\label{sec:settings}

\paragraph*{Datasets}

\begin{figure}[!h]
  \centering
  \begin{tabular}[c]{ll}
 \begin{subfigure}{0.35\columnwidth}
     \includegraphics[width=\columnwidth]{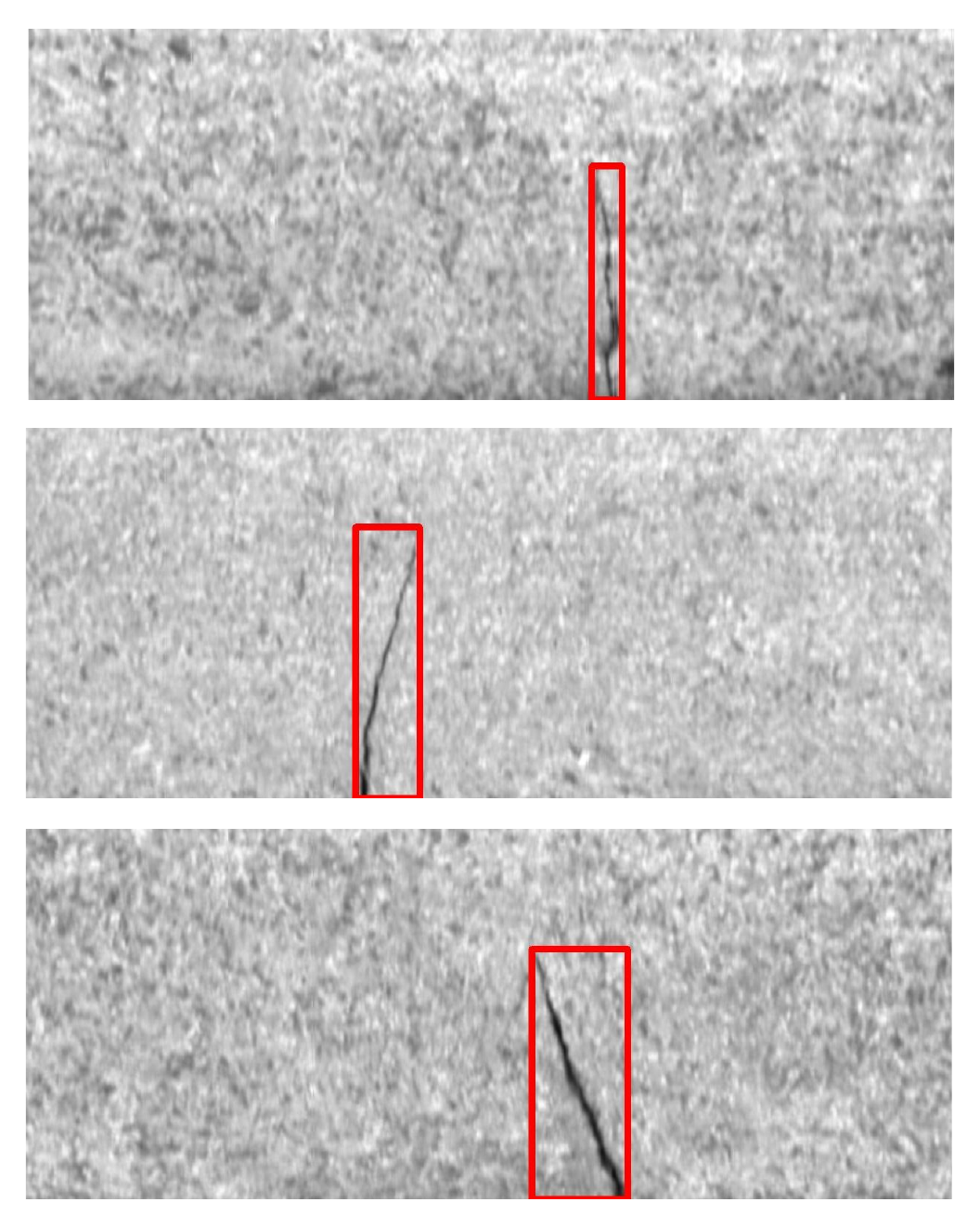}
     \vspace{-0.1cm}
     \caption{{\sf KSDD} data}
     \label{fig:ksdddata}
 \end{subfigure}&
 \begin{subfigure}{0.55\columnwidth}
     \includegraphics[width=\columnwidth]{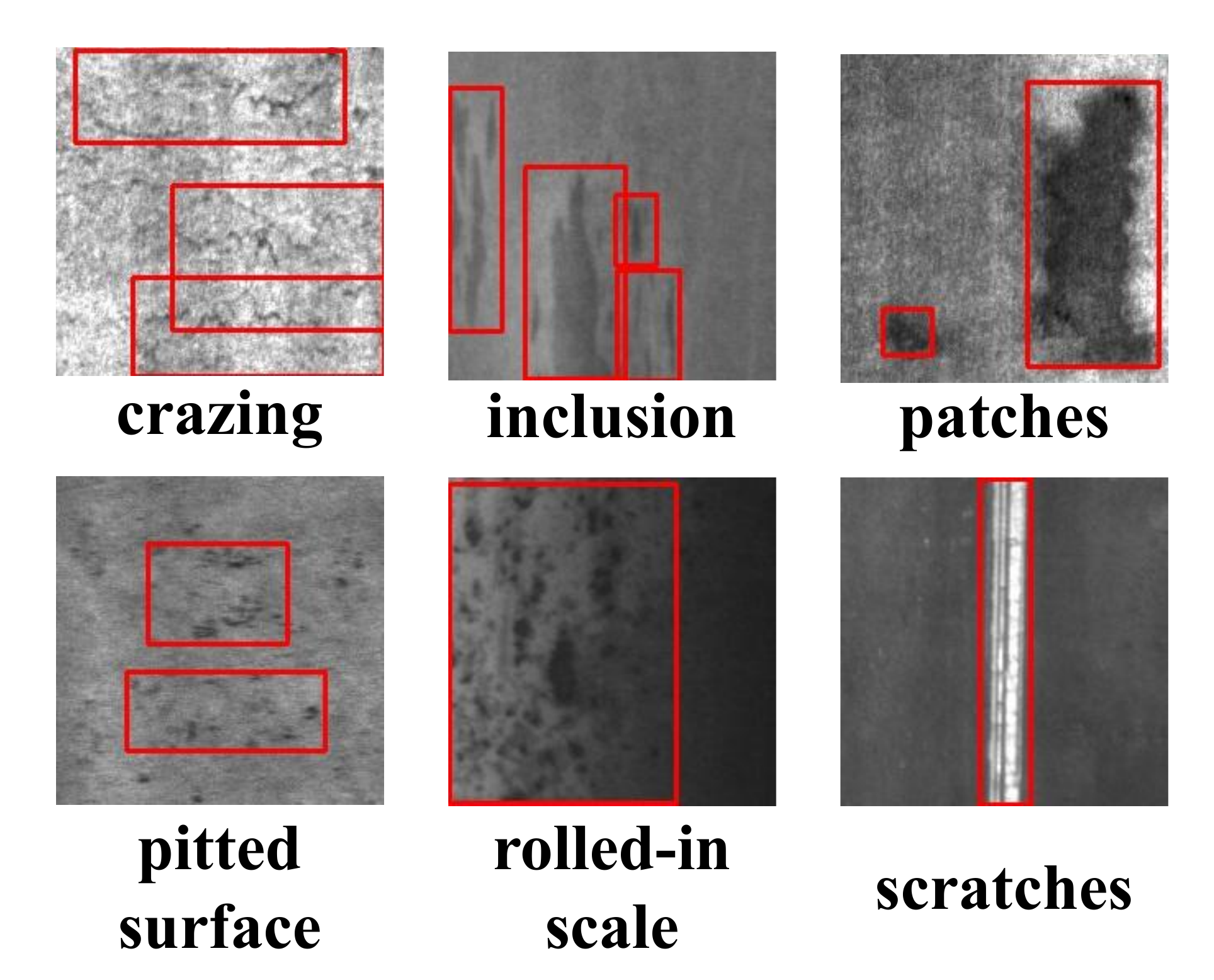}
     \vspace{-0.1cm}
     \caption{{\sf NEU} data}
     \label{fig:neudetdata}
 \end{subfigure}
 \end{tabular}
 \vspace{-0.3cm}
 \caption{Sample images in {\sf KSDD}~\protect\cite{Tabernik2019JIM} and {\sf NEU}~\protect\cite{He2019TIM} datasets where we highlight the defects with bounding boxes.}
 \label{fig:data}
 \vspace{-0.3cm}
\end{figure}

We use real datasets for \rv{{\em classification}} tasks. For each dataset, we construct a development set as described in Section~\ref{sec:crowdsourcing}. Table~\ref{tbl:datasets} summarizes the datasets with other experimental details, and \rv{Figures~\ref{fig:productdata} and ~\ref{fig:data}} shows samples of them. We note that the results in Section~\ref{sec:accuracyexperiments} are obtained by varying the size of the development set, and the rest of the experiments utilize the same size as described in Table~\ref{tbl:datasets}. 
For each dataset, we have a gold standard of labels. Hence, we are able to compute the accuracy of the labeling on separate test data.

\squishlist
    \item The Kolektor Surface-Defect Dataset ({\sf KSDD}~\cite{Tabernik2019JIM}) is constructed from images of electrical commutators that were provided and annotated by the Kolektor Group. There is only one type of defect -- cracks -- but each one varies significantly in shape.
   \item The {\sf Product} dataset (Figure~\ref{fig:productdata}) is proprietary and obtained through a collaboration with a manufacturing company. Each product has a circular shape where different strips are spread into rectangular shapes. There are three types of defects: scratches, bubbles, and stampings, which occur in different strips. The scratches vary in length and direction. The bubbles are more uniform, but have small sizes. The stampings are small and appear in fixed positions. We divide the dataset into three, as if there is a separate dataset for each defect type.
  \item The Northeastern University Surface Defect Database ({\sf NEU}~\cite{He2019TIM}) contains images that are divided into 6 defect types of  surface defects of hot-rolled steel strips: rolled-in scale, patches, crazing, pitted surface, inclusion, and scratches. Compared to the other datasets, these defects take larger portions of the images. Since there are no images without defects, we solve the different task of multi-class classification where the goal is to determine which defect is present.
\squishend


\paragraph*{GAN-based Augmentation} We provide more details for Section~\ref{subsec:ganaugmentation}. For all datasets, the input random noise vector has a size of 100, the learning rates of the generator and discriminator are both $1\mathrm{e}{-4}$, and the number of epochs is about 1K. We fit patterns to a square shape where the width and height are set to 100 or the averaged value of all widths and heights of patterns, whichever is smaller.

\paragraph*{Labeler Tuning} We use an L-BFGS optimizer~\cite{DBLP:journals/mp/LiuN89}, which provides stable training on small data, with a $1\mathrm{e}{-5}$ learning rate. We use $k$-fold cross validation where each fold has at least 20 examples per class and early stopping in order to compare the accuracies of \rv{candidate models} before they overfit. 

\paragraph*{Accuracy Measure} We use the $F_1$ score, which is the harmonic mean between precision and recall. Suppose that the set of true defects is $D$ while the set of predictions is $P$. Then the precision $Pr = \frac{|D \cap P|}{|P|}$, recall $Re = \frac{|D \cap P|}{|D|}$, and $F_1 = \frac{2 \times Pr \times Re}{Pr + Re}$. While there are other possible measures like ROC-AUC, $F_1$ is known to be more suitable for data where the labels are imbalanced~\cite{10.1007/978-3-642-04962-0_53} as in most of our settings.

\paragraph*{Systems Compared} \rv{We compare \ig{} with other image labeling systems and self-learning baselines that train CNN models on available labeled data.}

\snuba{}~\cite{DBLP:journals/pvldb/VarmaR18} automates the process of labeling function (LF) construction by starting from a set of primitives that are analogous to our FGFs and iteratively selecting subsets of them to train heuristic models, which becomes the LFs. Each iteration involves comparing models trained on all possible subsets of the primitives up to a certain size. Finally, the LFs are combined into a generative model. We faithfully implement \snuba{} and use our crowdsourced and augmented patterns for generating primitives, in order to be favorable to \snuba{}. However, adding more patterns quickly slows down \snuba{} as its runtime is exponential to the number of patterns.

We also compare with \goggles{}~\cite{DBLP:journals/corr/abs-1903-04552}, which takes the labeling approach of not using crowdsourcing. However, it relies on the fact that there is a pre-trained model and extracts semantic prototypes of images where each prototype represents the part of an image where the pre-trained model is activated the most. Each image is assumed to have one object, and \goggles{} clusters similar images for unsupervised learning. In our experiments, we use the opensourced code of \goggles{}. 

Finally, we compare \ig{} with self-learning~\cite{DBLP:journals/kais/TrigueroGH15} baselines that train CNN models on the development set using cross validation and use them to label the rest of the images. \rv{Note that when we compare \ig{} with a CNN model, we are mainly comparing their feature generation abilities. Inspector Gadget's features are the pattern similarities while for VGG's features are produced at the end of the convolutional layers. In both cases, the features go through a final fully-connected layer (i.e., MLP).} To make a fair comparison, we experiment with both heavy and light-weight CNN models. For the heavy model, we use VGG-19~\cite{DBLP:journals/corr/SimonyanZ14a}, which is widely used in the literature. For the light-weight model, we use MobileNetV2~\cite{DBLP:conf/cvpr/SandlerHZZC18}, which is designed to train efficiently in a mobile setting, but nearly has the performance of heavy CNNs. We also make a comparison with VGG-19 whose weights are pre-trained on ImageNet~\cite{DBLP:conf/cvpr/DengDSLL009} \rv{and fine-tuned on each dataset}. 
\ifthenelse{\boolean{techreport}}{
\tr{An alternative approach is to pre-train VGG-19 on other datasets in Table~\ref{tbl:datasets} instead of ImageNet. To see which approach is better, we compare the transfer learning results in Table~\ref{tbl:pretrain}. We observe that combining VGG-19 with ImageNet outperforms the other scenarios on all target datasets and thus use ImageNet to be favorable to transfer learning.}
}{
\rv{In our technical report~\cite{inspectorgadgettr}, we also considered variants of this transfer learning: 
using one of the other four datasets instead of ImageNet. However, our experiments show that combining VGG-19 with ImageNet performs the best.}}
In addition, we use preprocessing techniques on images that are favorable for the baselines. For example, the images from the {\sc Product} dataset are long rectangles, so we split each image in half and stack them on top of each other to make them more square-like, which is advantageous for CNNs.

\ifthenelse{\boolean{techreport}}{

\begin{table}[t]
  \caption{\tr{Comparison of VGG-19 F1 scores when pre-trained on various datasets.}}
  \vspace{-0.1cm}
  \begin{tabular}{@{\hspace{1pt}}l@{\hspace{3pt}}c@{\hspace{3pt}}c@{\hspace{3pt}}c@{\hspace{4pt}}c@{\hspace{4pt}}c@{\hspace{1pt}}}
    \toprule
     & \multicolumn{5}{c}{\bf Pre-trained on } \\
    \cmidrule(r){2-6}
    {\bf Target}& {\sf Product} & {\sf Product} & {\sf Product} &  & \\
  {\bf Dataset} & {\sf (scratch)}& {\sf (bubble)} & {\sf (stamping)} & {\sf KSDD} & {\sf ImageNet}\\
    \midrule
    {\sf Product (sc)} & x & 0.942 & 0.958 & 0.964 & \bf 0.972\\
    {\sf Product (bu)} & 0.535 & x & 0.531 & 0.453 & \bf 0.913 \\
    {\sf Product (st)} & 0.798 & 0.810 & x & 0.781 & \bf 0.900 \\
    {\sf KSDD} & 0.683 & 0.093 & 0.112 & x & \bf0.897 \\
    \bottomrule
  \end{tabular}
  \label{tbl:pretrain}
  \vspace{-0.2cm}
\end{table}

}{}

\begin{figure*}[t]
\vspace{-0.1cm}
  \centering
  \begin{subfigure}{0.32\textwidth}
     \includegraphics[width=\columnwidth]{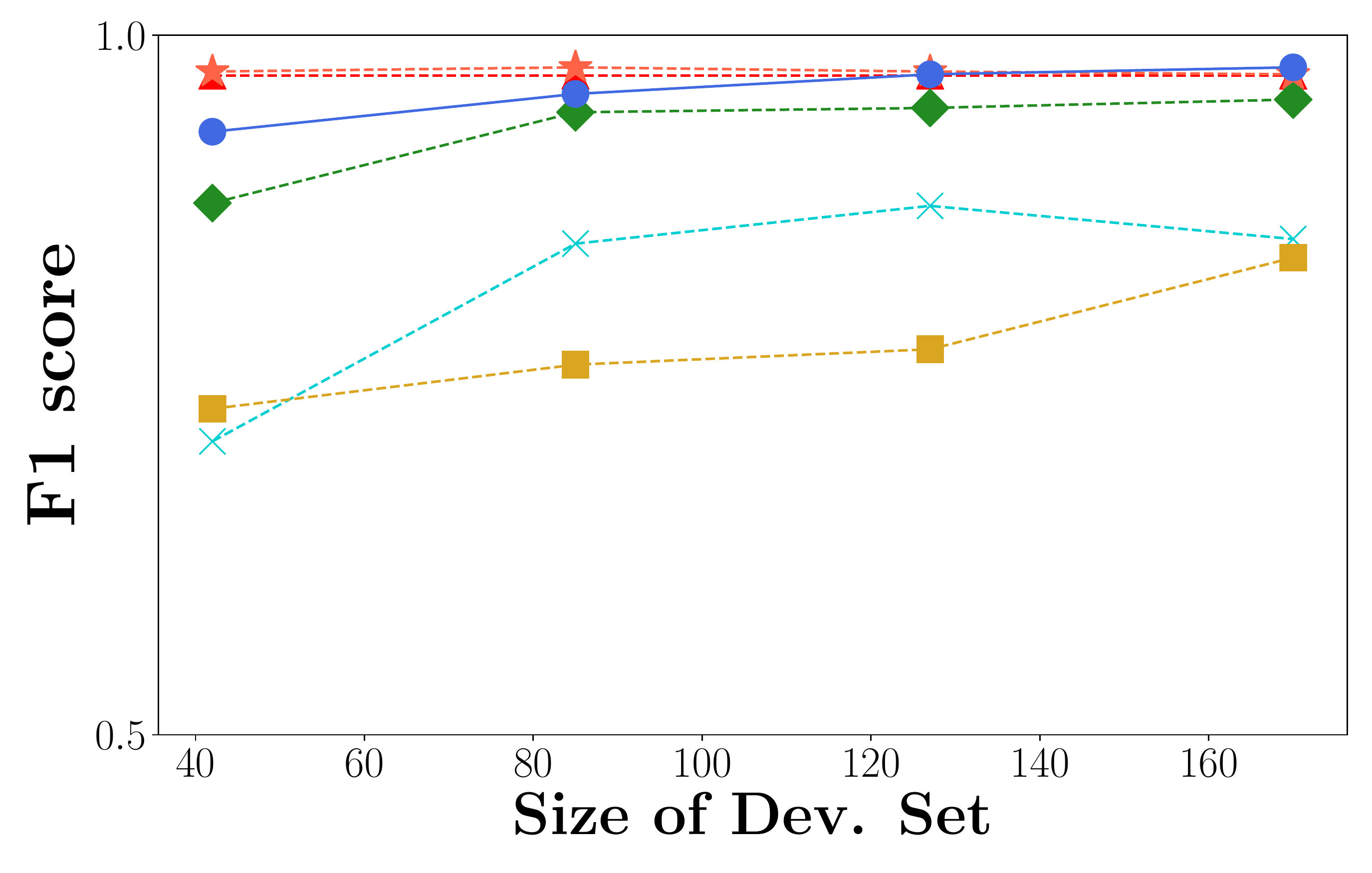}
     \caption{{\sf Product (scratch)}}
     \label{fig:ignversuscnn-scratch}
  \end{subfigure}
  \begin{subfigure}{0.32\textwidth}
     \includegraphics[width=\columnwidth]{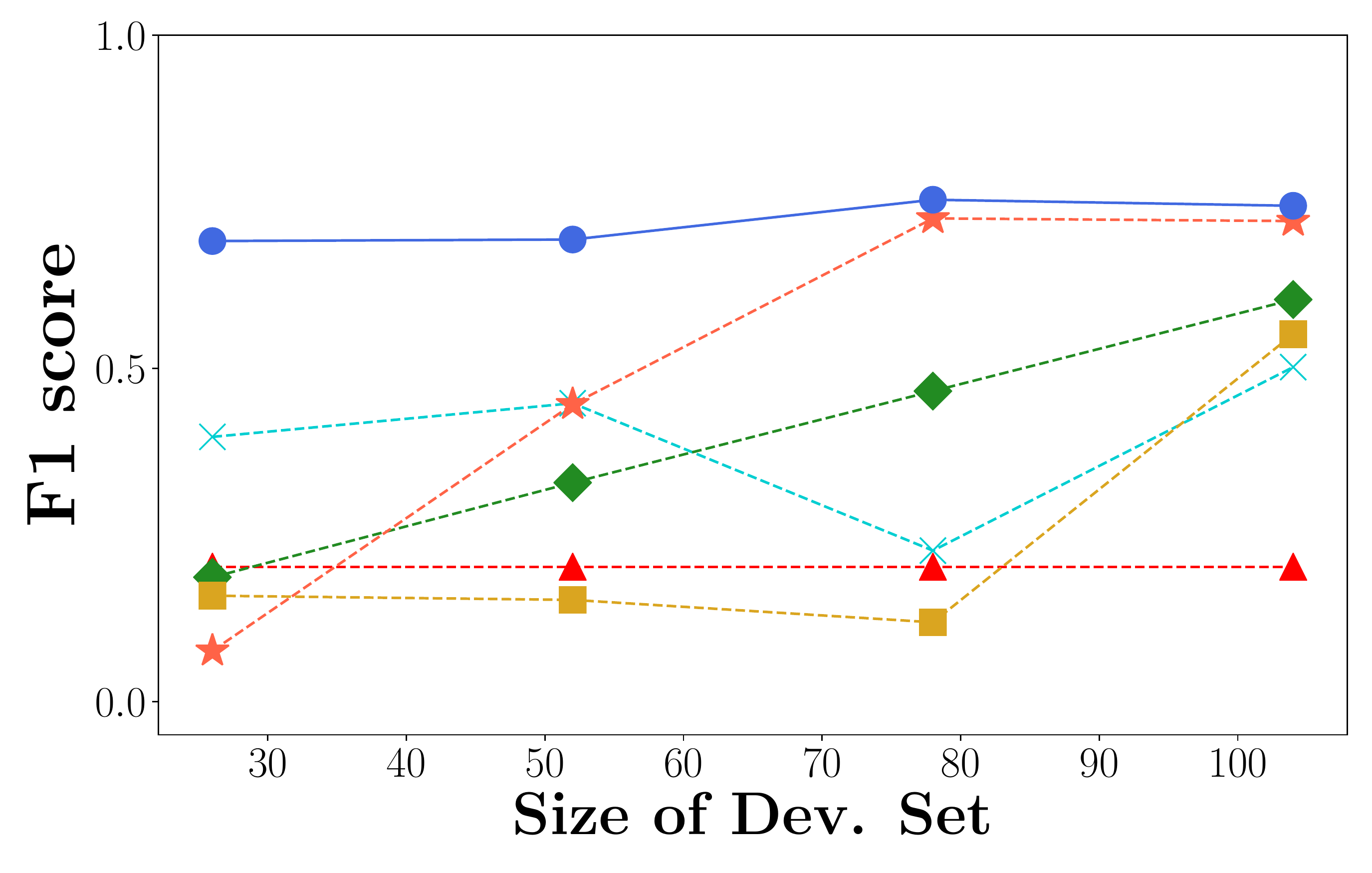}
     \caption{{\sf Product (bubble)}}
     \label{fig:ignversuscnn-bubble}
  \end{subfigure} 
  \begin{subfigure}{0.32\textwidth}
     \includegraphics[width=\columnwidth]{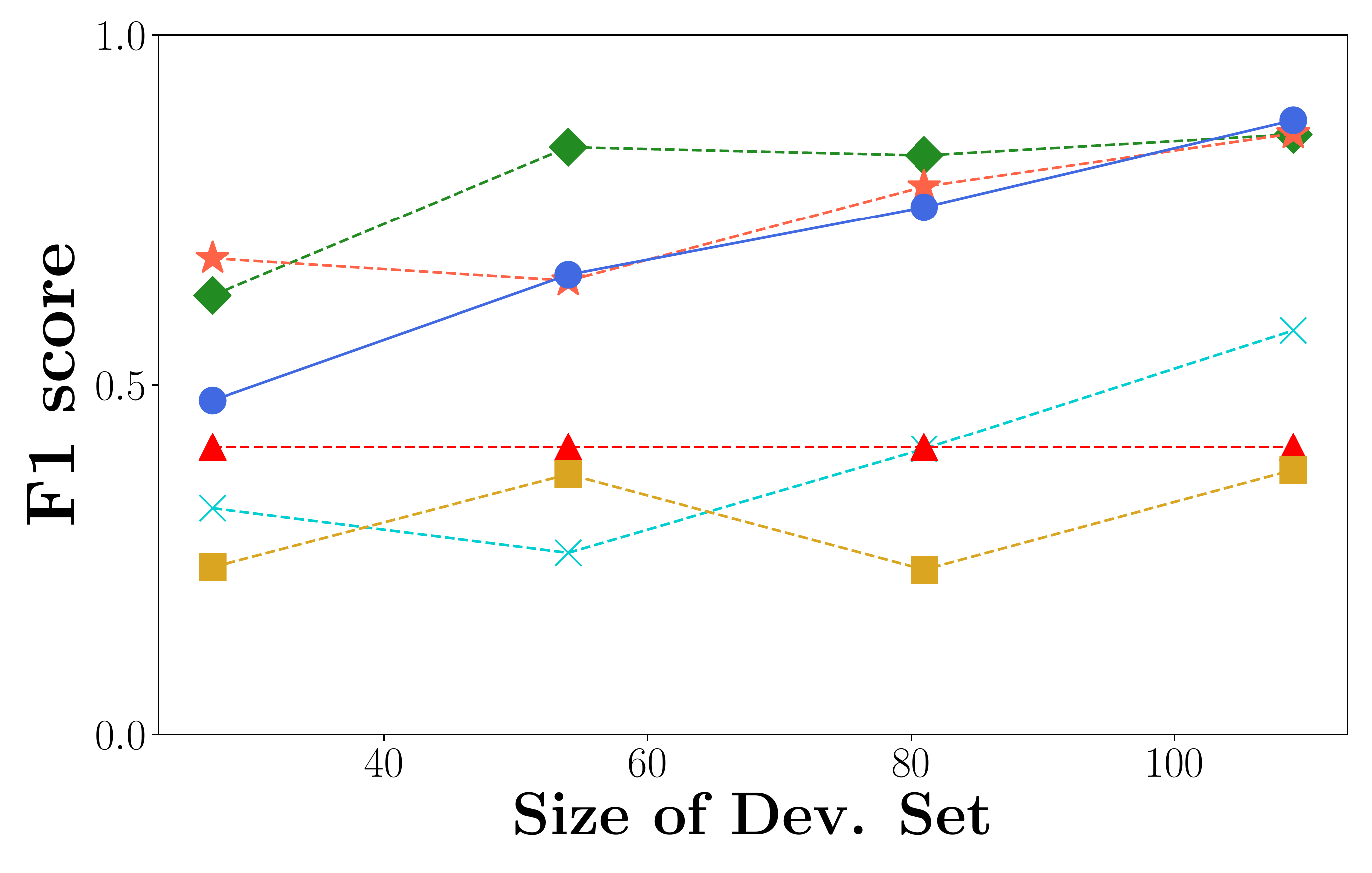}
     \caption{{\sf Product (stamping)}}
     \label{fig:ignversuscnn-stamping}
  \end{subfigure}
  \begin{subfigure}{0.32\textwidth}
     \includegraphics[width=\columnwidth]{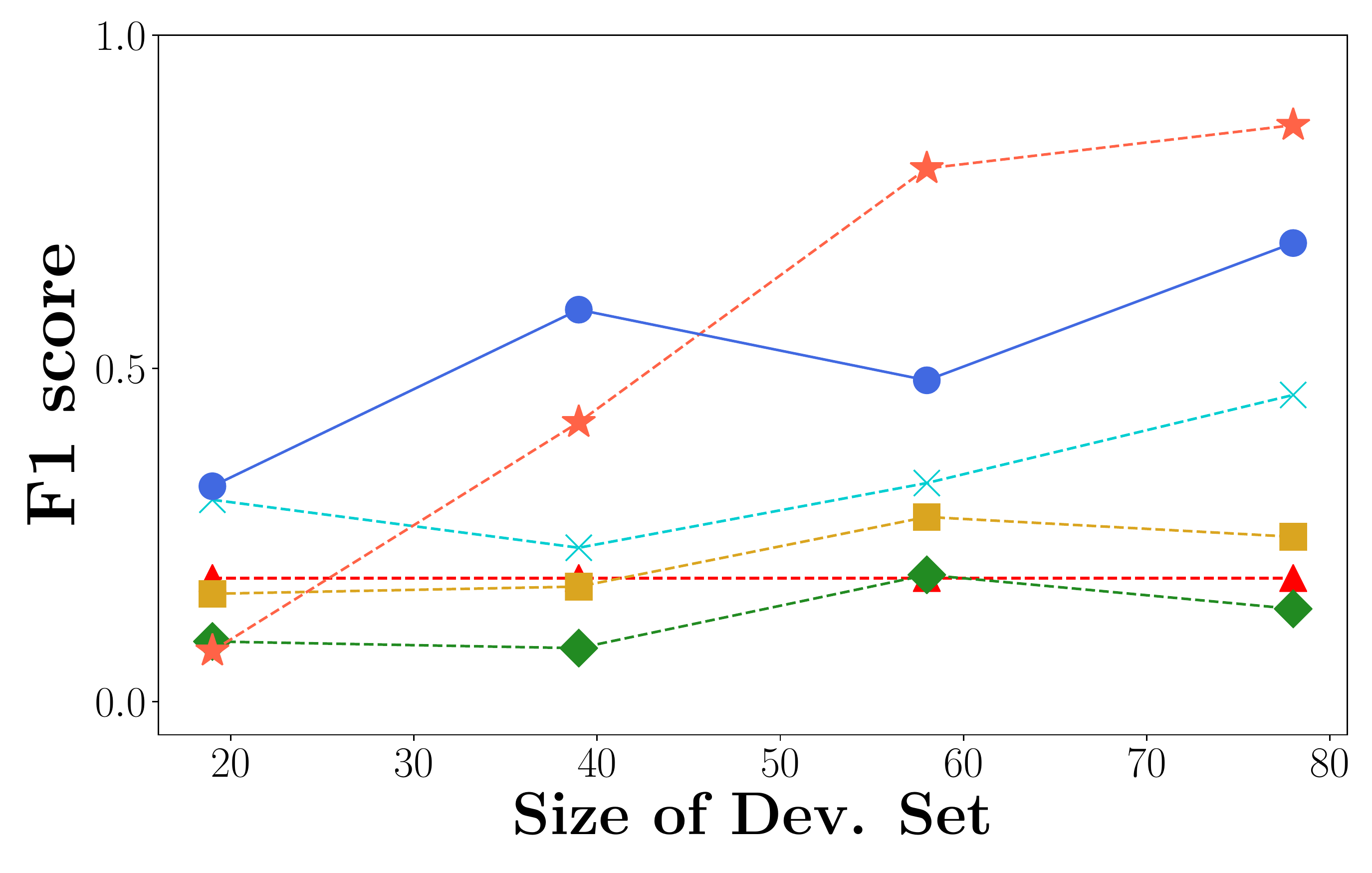}
     \caption{{\sf KSDD}~\cite{Tabernik2019JIM}}
     \label{fig:ignversuscnn-ksdd}
  \end{subfigure} 
  \begin{subfigure}{0.32\textwidth}
     \includegraphics[width=\columnwidth]{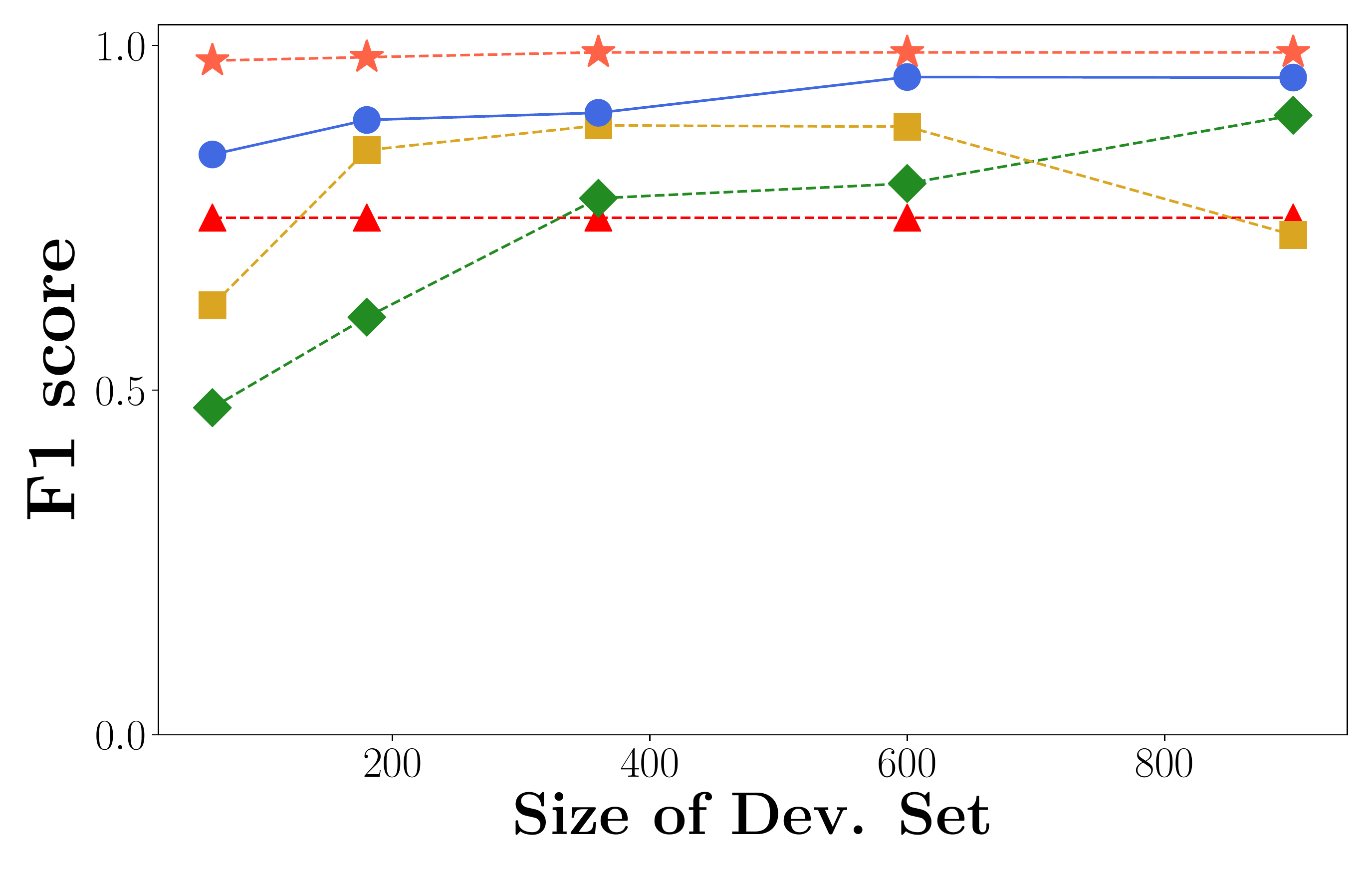}
     \caption{{\sf NEU}~\cite{He2019TIM}}
     \label{fig:ignversuscnn-neu}
  \end{subfigure}
  \begin{subfigure}{0.32\textwidth}
     {\hspace{25pt}}
     {\vspace{10pt}}
     \includegraphics[width=0.8\columnwidth]{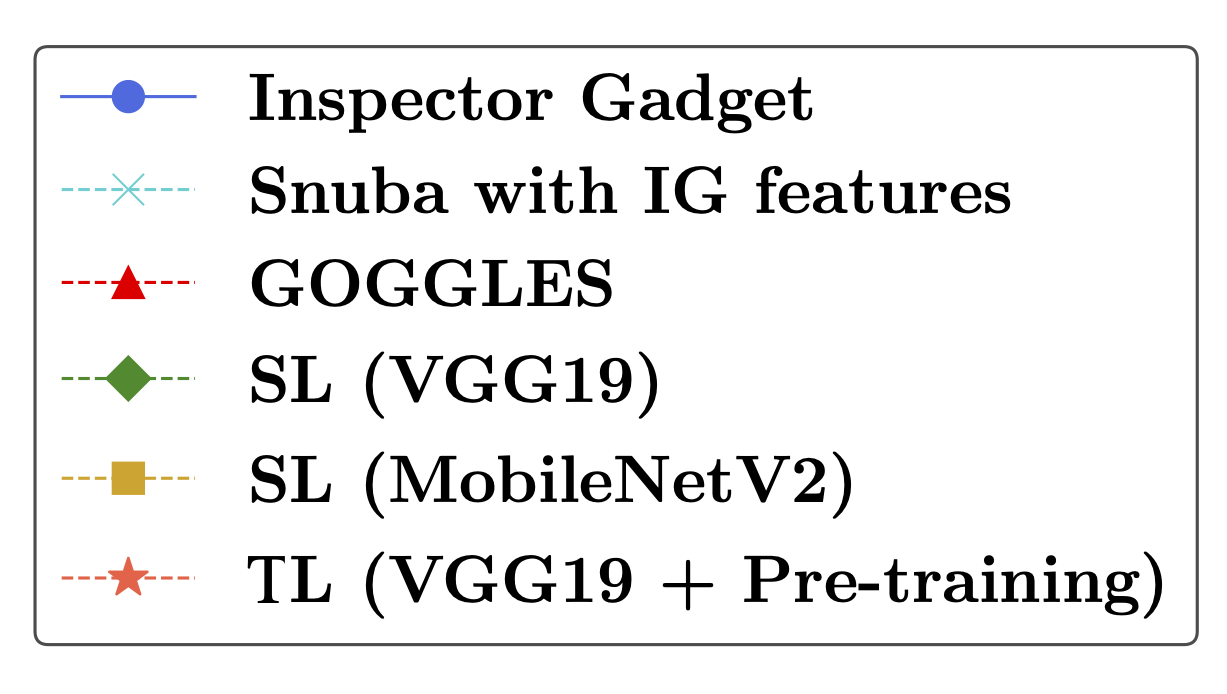}
  \end{subfigure} 
  \vspace{-0.25cm}
     \caption{Weak label accuracy comparison between \ig{}, \snuba{}~\protect\cite{DBLP:journals/pvldb/VarmaR18}, \goggles{}~\protect\cite{DBLP:journals/corr/abs-1903-04552}, the self-learning baselines (SL) using VGG-19~\protect\cite{DBLP:journals/corr/SimonyanZ14a} and MobileNetV2~\protect\cite{DBLP:conf/cvpr/SandlerHZZC18}, and transfer learning baseline (TL) on different sizes of development set. Among the models that are {\em not pre-trained}, \ig{} performs either the best or second-best in all figures.
     }
 \label{fig:igversuscnn}
 \vspace{-0.05cm}
\end{figure*}

\subsection{Weak Label Accuracy}
\label{sec:accuracyexperiments}


\rv{We compare the weak label accuracy of \ig{} with the other methods by increasing the development set size and observing the $F_1$ scores in Figure~\ref{fig:igversuscnn}.} To clearly show how \ig{} compares with other methods, we use a solid line to draw its plot while using dotted lines for the rest. Among the models that are {\em not pre-trained} (i.e., ignore ``TL (VGG19 + Pre-training)'' for now), we observe that \ig{} performs best overall because it is either the best or second-best method in all figures.
This result is important because industrial images have various defect types that must all be identified correctly. For {\sf KSDD} (Figure~\ref{fig:ignversuscnn-ksdd}), \ig{} performs the best because the pattern augmentation helps \ig{} find more variations of cracks (see Section~\ref{sec:dataaugmentationexperiments}). For {\sf Product} (Figures~\ref{fig:ignversuscnn-scratch}--\ref{fig:ignversuscnn-stamping}), \ig{} consistently performs the first or second best despite the different characteristics of the defects. For {\sf NEU} (Figure~\ref{fig:ignversuscnn-neu}), \ig{} ranks first for the multi-class classification.

We explain the performances of other methods. \snuba{} consistently has a lower $F_1$ than \ig{} possibly because the number of patterns is too large to handle. Instead of considering all combinations of patterns and training heuristic models, \ig{}'s approach of training the labeler works better for our experiments. \goggles{} does not \rv{use gold labels for training} and thus has a constant accuracy. In Figure~\ref{fig:ignversuscnn-scratch}, \goggles{} has a high $F_1$ because the defect sizes are large, and the pre-trained VGG-16 is effective in identifying them as objects. For the other figures, however, \goggles{} does not perform as well because the defect sizes are small and difficult to identify as objects. VGG-19 without pre-training (``SL (VGG19)'') only performs the best in Figure~\ref{fig:ignversuscnn-stamping} where CNN models are very good at detecting stamping defects because they appear in a fixed location on the images. For other figures, VGG-19 performs poorly because there is not enough labeled data. MobileNetV2 does not perform well in any of the figures. \rv{Finally, the transfer learning method (``TL (VGG19 + Pre-training)'') shows a performance comparable to \ig{}. In particular, transfer learning performs better in Figures~\ref{fig:ignversuscnn-scratch}, \ref{fig:ignversuscnn-stamping}, and \ref{fig:ignversuscnn-neu}, while \ig{} performs better in Figures~\ref{fig:ignversuscnn-bubble} and \ref{fig:ignversuscnn-ksdd} (for small dev. set sizes). While we do not claim that \ig{} outperforms transfer learning models, it is thus an attractive option for certain types of defects (e.g., bubbles) and small dev.\@ set sizes.} 

\subsection{Crowdsourcing Workflow}
\label{sec:workflowexperiments}

We evaluate how effectively we can use the crowd to label and identify patterns using the {\sf Product} datasets. Table~\ref{tbl:workflow} compares the full crowdsourcing workflow in \ig{} with two variants: (1) a workflow that does not average the patterns at all and (2) a workflow that does average the patterns, but still does not perform peer reviews. \rv{We evaluate each scenario without using pattern augmentation.} 
As a result, the full workflow clearly performs the best for the {\sf Product} (scratch) and {\sf Product} (stamping) datasets. For the {\sf Product} (bubble) dataset, the workflow that does not combine patterns has a better average $F_1$, but the accuracies vary among different workers. Instead, it is better to use the stable full workflow without the variance.

\begin{table}[t]
  \vspace{-0.25cm}
  \caption{Crowdsourcing workflow ablation results}
  \vspace{-0.2cm}
  \begin{tabular}{@{\hspace{1pt}}l@{\hspace{3pt}}c@{\hspace{3pt}}c@{\hspace{3pt}}c@{\hspace{1pt}}}
    \toprule
     & \multicolumn{3}{c}{\bf $F_1$ scores } \\
    \cmidrule(r){2-4}
    & \bf  No  avg. & \bf No & \bf Full \\
  {\bf Dataset} & \bf  (${\pm std/2}$)& \bf peer review & \bf workflow\\
    \midrule
    {\sf Product (scratch)} & 0.940 (${\pm 0.005}$) & 0.952 & 0.960\\
    {\sf Product (bubble)} & 0.616 (${\pm 0.045}$) & 0.525 & 0.605 \\
    {\sf Product (stamping)} & 0.299 (${\pm 0.142}$) & 0.543 & 0.595 \\
    \bottomrule
  \end{tabular}
  \label{tbl:workflow}
  \vspace{-0.25cm}
\end{table}

\begin{table}[t]
  \caption{Pattern augmentation impact on \ig{}. For each dataset, we highlight the highest $F_1$ score. }
  \vspace{-0.2cm}
  \begin{tabular}{ccccc}
    \toprule
    \bf \multirow{2}{*}{Dataset} & \bf No & \bf Policy & \bf GAN & \bf Using \\
    & \bf Aug. & \bf Based & \bf Based & \bf Both \\
    \midrule
    \multirow{1}{*}{\sf KSDD}~\cite{Tabernik2019JIM} & 0.415 & 0.578 & 0.509 & \bf 0.688 \\
    \multirow{1}{*}{{\sf Product (scratch)}} & 0.958 & 0.965 & 0.962 & \bf 0.979 \\
    \multirow{1}{*}{{\sf Product (bubble)}} & 0.617 & 0.702 & \bf 0.715 & 0.701 \\
    \multirow{1}{*}{{\sf Product (stamping)}} & 0.700 & 0.700  & 0.765 & \bf 0.859 \\
    \multirow{1}{*}{{\sf NEU}~\cite{He2019TIM}} & 0.936 & \bf 0.954 & 0.930 & \bf 0.954 \\
    \bottomrule
  \end{tabular}
  
  \label{tbl:dataaugmentation}
  \vspace{-0.4cm}
\end{table}

\subsection{Pattern Augmentation}
\label{sec:dataaugmentationexperiments}

We evaluate how augmented patterns help improve the weak label $F_1$ score of \ig{}. Table~\ref{tbl:dataaugmentation} shows the impact of the GAN-based and policy-based augmentation on the five datasets. 
\ifthenelse{\boolean{techreport}}{
\tr{
Figure~\ref{fig:augmentation} shows how adding patterns impacts the $F_1$ score for the {\sf Product} (bubble) dataset. While adding more patterns helps to a certain extent, it has diminishing returns afterwards. The results for the other datasets are similar, although sometimes noisier. The best number of augmented patterns differs per dataset, but falls in the range of 100--500.
}
}{
For each augmentation, we add 100--500 patterns, which results in the best $F_1$ improvements (see our technical report~\cite{inspectorgadgettr} for an empirical analysis). 
}
When using both methods, we simply combine the patterns from each augmentation. As a result, while each augmentation helps improve $F_1$, using both of them usually gives the best results. While adding more patterns thus helps to a certain extent, it does has diminishing returns afterwards.

\rv{Pattern augmentation is an important way to solve class imbalance where there are very few defects compared to non-defects. Among the five datasets, {\sf KSDD}, {\sf Product (bubble)}, and {\sf Product (stamping)} have relatively few numbers of defects, and they benefit the most from pattern augmentation where the performance lift compared to no augmentation ranges from 0.10--0.27.}

\ifthenelse{\boolean{techreport}}{

\begin{figure}[h]
  \centering
     \includegraphics[width=0.8\columnwidth]{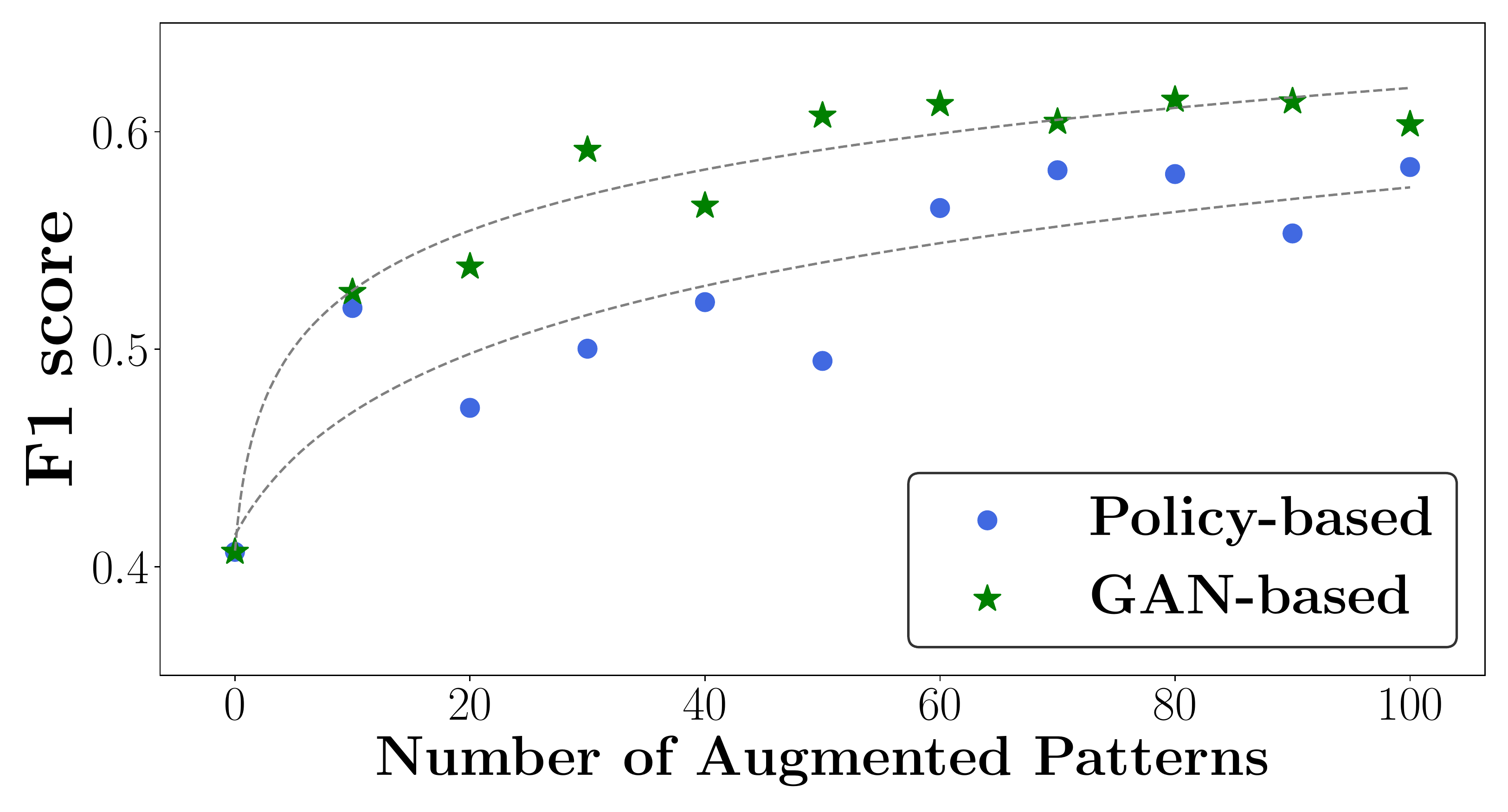}
     \caption{\tr{Policy-based and GAN-based augmentation results on the {\sf Product} (stamping) dataset.}} 
 \label{fig:augmentation}
\end{figure}

}{}

\subsection{Model Tuning}
\label{sec:modeltuningexperiments}

\begin{figure}[t]
\vspace{-0.3cm}
\centering
     \includegraphics[width=0.9\columnwidth]{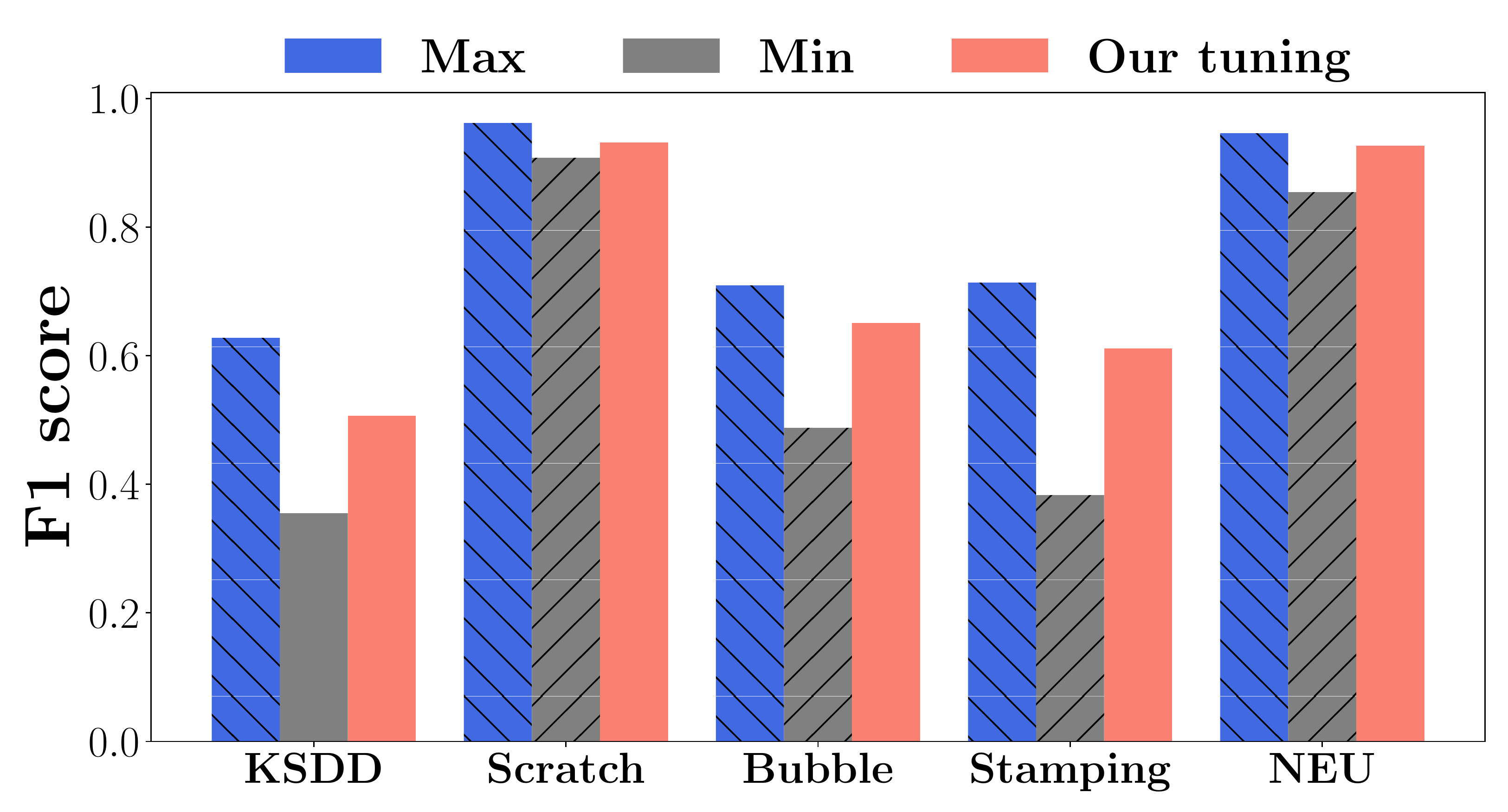}
\vspace{-0.3cm}       
\caption{The variations in $F_1$ scores when tuning the MLP model hyper-parameters.}
\label{fig:augmentationandtuning}
\vspace{-0.8cm}
\end{figure}

We evaluate the impact of model tuning on accuracy described in Section~\ref{sec:modeltraining} as shown in Figure~\ref{fig:augmentationandtuning}. We use an MLP with 1 to 3 hidden layers and varied the number of nodes per hidden layer to be one of \{$2^n | n = 1 \ldots m \text{ and } 2^{m-1} \leq I \leq 2^m$\} where $I$ is the number of input nodes. For each dataset, we first obtain the maximum and minimum possible $F_1$ scores by evaluating all the tuned models we considered directly on the test data. \rv{Then, we compare these results with the (test data) $F_1$ score of the actual model that \ig{} selected using the development set. } 
We observe that the model tuning in \ig{} can indeed improve the model accuracy, close to the maximum possible value.

\subsection{End Model Accuracy}
\label{sec:endmodelaccuracy}

We now address the issue of whether the weak labels are actually helpful for training the end discriminative model. We compare the $F_1$ score of this end model with the same model that is trained on the development set. For the discriminative model, we use VGG-19~\cite{DBLP:journals/corr/SimonyanZ14a} for the binary classification tasks on {\sf KSDD} and {\sf Product}, and ResNet50~\cite{DBLP:conf/cvpr/HeZRS16} for the multi-class task on {\sf NEU}. We can use other discriminative models that have higher absolute $F_1$, but the point is to show the {\em relative} $F_1$ improvements when using weak labels. Table~\ref{tbl:endmodel} shows that the $F_1$ scores improve by 0.02--0.36. \rv{In addition, the results of ``Tip.\@ Pnt'' show that one needs to increase the sizes of the development sets by 1.874--7.565x for the discriminative models to obtain the same $F_1$ scores as \ig{}.}

\begin{table}[t]
  \ifthenelse{\boolean{techreport}}{\vspace{0.4cm}}
  
  \caption{$F_1$ scores of end models trained on the development set (Dev. Set) only or the development set combined with the weak labels produced by \ig{} (WL (IG)). \rv{{\bf Tip. Pnt} shows how larger the development set must be for the {\bf Dev. Set} approach to obtain the same $F_1$ score as {\bf WL (IG)}.}}
  
  {
  \vspace{-0.2cm}
  }
  \begin{tabular}{@{\hspace{6pt}}c@{\hspace{6pt}}c@{\hspace{6pt}}c@{\hspace{6pt}}c@{\hspace{6pt}}c@{\hspace{6pt}}}
    \toprule
    \bf Dataset & \bf End Model & \bf Dev. Set & \bf WL (IG) & \bf \rv{Tip. Pnt}\\
   \midrule
    \multirow{1}{*}{\sf KSDD}~\cite{Tabernik2019JIM} & VGG19 & 0.499 & 0.700 & \rv{$\times$3.248}\\
    \multirow{1}{*}{{\sf Product (sc)}}& VGG19 & 0.925 & 0.978 & \rv{$\times$4.374}\\
    \multirow{1}{*}{{\sf Product (bu)}}& VGG19 & 0.359 & 0.720 & \rv{$\times$6.041}\\
    \multirow{1}{*}{{\sf Product (st)}}& VGG19 & 0.782 & 0.876 & \rv{$\times$7.565}\\
    \multirow{1}{*}{{\sf NEU}~\cite{He2019TIM}}& ResNet50 & 0.953 & 0.970 & \rv{$\times$1.874}\\
    \bottomrule
  \end{tabular}
  \label{tbl:endmodel}
  \vspace{-0.2cm}
\end{table}

\subsection{Error Analysis}

We perform an error analysis on which cases \ig{} fails to make correct predicts for the five datasets based on manual investigation. We use the ground truth information for the analysis. Table~\ref{tbl:erroranalysis} shows that most common error is when certain defects do not match with the patterns, which can be improved by using better pattern augmentation and matching techniques. The next common case is when the data is noisy, which can be improved by cleaning the data. The last case is the most challenging where even humans have difficulty identifying the defects because they are not obvious (e.g., a near-invisible scratch).


\begin{table}[t]
  \centering
  \caption{Error analysis of \ig{}.}
  \vspace{-0.2cm}
  \begin{tabular}{@{\hspace{4pt}}c@{\hspace{4pt}}c@{\hspace{4pt}}c@{\hspace{4pt}}c@{\hspace{4pt}}}
    \toprule
     & \multicolumn{3}{c}{\bf Cause } \\
    \cmidrule(r){2-4}
    & \bf  Matching & \bf Noisy & \bf Difficult \\
  {\bf Dataset} & \bf failure & \bf data & \bf to humans \\
    \midrule
    \multirow{1}{*}{\sf KSDD}~\cite{Tabernik2019JIM} & 10 (52.6 \%) & 5 (26.3 \%) & 4 (21.1 \%) \\
    \multirow{1}{*}{\sf Product (scratch)} & 11 (36.7 \%) & 11 (36.7 \%) & 8 (26.6 \%)\\
    \multirow{1}{*}{\sf Product (bubble)} & 19 (45.2 \%) & 15 (35.7 \%) & 8 (19.1 \%) \\
    \multirow{1}{*}{\sf Product (stamping)} & 15 (45.5 \%) & 13 (39.4 \%) & 5 (15.1 \%) \\
    \multirow{1}{*}{\sf NEU}~\cite{He2019TIM} & 35 (63.6 \%) &  4 (7.3 \%) & 16 (29.1 \%) \\
    \bottomrule
  \end{tabular}
  \label{tbl:erroranalysis}
  \vspace{-0.3cm}
\end{table}

\section{Related Work}
\label{sec:relatedwork}


\paragraph*{Crowdsourcing for machine learning \rv{and databases}}

\rv{Using humans in advanced analytics is increasingly becoming mainstream~\cite{DBLP:conf/sigmod/XinMLMSP18} where the tasks include query processing~\cite{marcus2011crowdsourced}, entity matching~\cite{gokhale2014corleone}, active learning~\cite{mozafari2012active}, data labeling~\cite{haas2015clamshell}, and feature engineering~\cite{DBLP:conf/cscw/ChengB15}. \ig{} provides a new use case where the crowd identifies patterns.}


\paragraph*{Data Programming}

Data programming~\cite{DBLP:conf/nips/RatnerSWSR16} is a recent paradigm where workers program labeling functions (LFs), which are used to generate weak labels at scale. \snorkel{}~\cite{Ratner:2017:SFT:3035918.3056442,DBLP:conf/sigmod/BachRLLSXSRHAKR19} is a seminal system that demonstrates the practicality of data programming, and \snuba{}~\cite{DBLP:journals/pvldb/VarmaR18} extends it by automatically constructing LFs using primitives. In comparison, \ig{} does not assume any accuracy guarantees on the feature generation function and directly labels images without converting them to structured data.

Several systems have studied the problem of automating labeling function construction. CrowdGame~\cite{DBLP:conf/sigmod/LiuYFWLD19} proposes a method for constructing LFs for entity resolution on structured data. Adversarial data programming~\cite{DBLP:conf/cvpr/PalB18} proposes a GAN-based framework for labeling with LF results and claims to be better than \snorkel{}-based approaches. In comparison, \ig{} solves the different problem of partially analyzing large images.

\paragraph*{Automatic Image labeling}

There is a variety of general automatic image labeling techniques. Data augmentation~\cite{DBLP:journals/jbd/ShortenK19} is a general method to generate new labeled images. Generative adversarial networks (GANs)~\cite{DBLP:conf/nips/GoodfellowPMXWOCB14} have been proposed to generate fake, but realistic images based on existing images. Policies~\cite{DBLP:journals/corr/abs-1805-09501} were proposed to apply custom transformations on images as long as they remain realistic. Most of the existing work operate on the entire images. In comparison, \ig{} is efficient because it only needs to augment patterns, which are much smaller than the images. Label propagation techniques~\cite{DBLP:conf/wsdm/BuiRR18} organize images into a graph based on their similarities and then propagates existing labels of images to their most similar ones. In comparison, \ig{} is designed for images where only a small part of them are of interest while the main part may be nearly identical to other images, so we cannot utilize the similar method. There are also application-specific defect detection methods~\cite{song2013noise,He2019TIM,song2019semisupervised}, some of which are designed for the datasets we used. In comparison, \ig{} provides a general framework for image labeling. Recently, \goggles{}~\cite{DBLP:journals/corr/abs-1903-04552} is an image labeling system that relies on a pre-trained model \rv{to extract semantic prototypes of images. }
In comparison, \ig{} does not rely on pre-trained models and is more suitable for partially analyzing large images using human knowledge. \rv{Lastly, an interesting line of work is novel class detection~\cite{scheirer2012toward} where the goal is to identify unknown defects. While \ig{} assumes a fixed set of defects, it can be extended with these techniques.}


\section{Conclusion}

We proposed \ig{}, a scalable image labeling system for classification problems that effectively combines crowdsourcing, data augmentation, and data programming techniques. \ig{} targets applications in manufacturing where large industrial images are partially analyzed, and there are few or no labels to start with. Unlike existing data programming approaches that convert images to structured data beforehand \rv{
using object detection models}, \ig{} directly labels images by providing a crowdsourcing workflow to leverage human knowledge for identifying patterns of interest. The patterns are then augmented and matched with other images to generate similarity features for MLP model training. Our experiments show that \ig{} outperforms \rv{other image labeling methods (\snuba{}, \goggles{})} and self-learning baselines using CNNs without pre-training. 
We thus believe that \ig{} opens up a new class of problems to apply data programming. 

\section{Acknowledgments}

This work was supported by a Google AI Focused Research Award, by SK Telecom, and by the Engineering Research Center Program through the National Research Foundation of Korea (NRF) funded by the Korean Government MSIT (NRF-2018R1A5A1059921).


\bibliographystyle{ACM-Reference-Format}
\bibliography{main}

\end{document}